%% file: main_pami.tex
\documentclass[10pt,journal,compsoc]{IEEEtran}
\ifCLASSOPTIONcompsoc
  \usepackage[nocompress]{cite}
\else
  \usepackage{cite}
\fi

\ifCLASSINFOpdf
  \usepackage[pdftex]{graphicx}
\else
   \usepackage[dvips]{graphicx}
\fi

\usepackage{amsmath}
\usepackage{amsmath}
\usepackage{amsthm, amsfonts}
\usepackage{balance}
\usepackage{algorithm}
\usepackage{algorithmic}
\newtheorem{theorem}{Theorem}

\usepackage{longtable}
\usepackage{cite}
\usepackage{enumerate}
\usepackage{subfigure}
\usepackage{float}
\usepackage{multicol}
\usepackage{color}
\usepackage{array}
\usepackage{booktabs}
\usepackage{multirow}
\usepackage{xcolor}
\usepackage{url}
\usepackage[nobiblatex]{xurl}
\usepackage{ragged2e}
\usepackage{enumitem}
\usepackage{amssymb}
\usepackage{amsthm}

\begin{document}
%
\title{ControlVAE: Tuning, Analytical Properties, and Performance Analysis}
%
%
%
%
\author{Huajie~Shao$^*$,
		Zhisheng~Xiao$^*$,
        Shuochao Yao,
        Dachun~Sun,
         Aston Zhang, \\
        Shengzhong Liu,
        Tarek Abdelzaher,~\IEEEmembership{Fellow,~ACM}
\IEEEcompsocitemizethanks{\IEEEcompsocthanksitem Huajie Shao, Shuochao, Dachun Sun, Shengzhong Liu, Tianshi Wang and Tarek Abdelzaher are with Department of Computer Science, University of Illinois at Urbana-Champaign, Urbana, IL 61801.   \protect\\
E-mail: \{hshao5, syao9, dsun18, sl29, tianshi3, zaher\}@illinois.edu
\IEEEcompsocthanksitem Zhisheng Xiao is with Committee on Computational and Applied Mathematics at the University of Chicago, Chicago 60637. \protect\\
E-mail: zxiao@uchicago.edu
\IEEEcompsocthanksitem Aston Zhang is with Amazon Web Services, CA.\protect\\
E-mail: astonz@amazon.com
}
\thanks{$^*$The first two authors contribute equally.}
\thanks{Manuscript received **, 2020;}}

%
%

\markboth{IEEE Transactions on Pattern Analysis and Machine Intelligence,~Vol.~**, No.~*, July~2020}%
{Huajie Shao \MakeLowercase{\textit{et al.}}: Unsupervised Truth Discovery with Multi-modal Data in Social Sensing}
%



\IEEEtitleabstractindextext{%
\justify
\input{Abstract}

}

\maketitle

\IEEEdisplaynontitleabstractindextext

%
\IEEEpeerreviewmaketitle


\input{introduction}

\input{preliminary}
\input{model}
\input{Evaluation}

\input{ablation}
\input{relatedwork}

\input{conclusion}

\section*{Acknowledgements}
Research reported in this paper was sponsored in part by DARPA award W911NF-17-C-0099, DTRA award HDTRA1-18-1-0026, and the Army Research Laboratory under Cooperative Agreements W911NF-09-2-0053 and W911NF-17-2-0196.

\ifCLASSOPTIONcaptionsoff
  \newpage
\fi



%
\bibliographystyle{IEEEtran}
\bibliography{references}
\clearpage
\input{appendix}

\end{document}

%% file: abstract.tex
\begin{abstract}
This paper reviews the novel concept of controllable variational autoencoder (ControlVAE), discusses its parameter tuning to meet application needs, derives its key analytic properties, and offers useful extensions and applications. ControlVAE is a new variational autoencoder (VAE) framework that combines the automatic control theory with the basic VAE to stabilize the KL-divergence of VAE models to a specified value. It leverages a non-linear PI controller, a variant of the proportional-integral-derivative (PID) control, to dynamically tune the weight of the KL-divergence term in the evidence lower bound (ELBO) using the output KL-divergence as feedback. This allows us to precisely control the KL-divergence to a desired value (set point), which is effective in avoiding posterior collapse and learning disentangled representations. In order to improve the ELBO over the regular VAE, we provide simplified theoretical analysis to inform setting the set point of KL-divergence for ControlVAE. We observe that compared to other methods that seek to balance the two terms in VAE's objective, ControlVAE leads to better learning dynamics. In particular, it can achieve a good trade-off between reconstruction quality and KL-divergence. We evaluate the proposed method on three tasks: image generation, language modeling and disentangled representation learning. The results show that ControlVAE can achieve much better reconstruction quality than the other methods for comparable disentanglement. On the language modeling task, ControlVAE can avoid posterior collapse (KL vanishing) and improve the diversity of generated text. Moreover, our method can change the optimization trajectory, improving the ELBO and the reconstruction quality for image generation.
\end{abstract}

%% file: introduction.tex
\section{Introduction}
\label{sec:introduction}

A Variational Auto-encoder~\cite{kingma2013auto, rezende2014stochastic} (VAE) is a type of generative model that has been widely used in a variety of applications, such as image generation~\cite{yan2016attribute2image,liu2017unsupervised}, dialog generation~\cite{zhang2020dive,wang2019topic,hu2017toward}, and disentangled representation learning~\cite{higgins2017beta,kim2018disentangling}. A VAE is composed of an encoder that maps input data to a distribution of latent variables, and a decoder that maps a sampled latent code back to the data space. The encoder and decoder are trained jointly by minimizing the reconstruction loss between input data and output of the decoder, plus the KL-divergence between the latent distribution and a pre-defined prior, such as the unit Gaussian.

The two terms in the objective have contrasting effects: the reconstruction term improves the reconstruction quality while neglecting the structure of the latent space; the KL-divergence term regularizes the latent space, possibly at the cost
of some overlapping between latent variables, hence resulting in a more noisy encoding. In practice, depending on the tasks, we often want to find a desirable position in such a trade-off. For example, in text or image generation, the goal is to generate \textit{diverse} and \textit{new} text or images, as opposed to reproducing one of the training samples. If KL-divergence is too low, the output samples have very limited diversity (known as the KL-vanishing or posterior collapse problem ~\cite{bowman2015generating}). To increase output diversity, it becomes advantageous to artificially {\em increase\/} the KL-divergence. Conversely, in disentangled representation learning~\cite{denton2017unsupervised}, we want to ensure independence among latent variables. In this context, artificially {\em decreasing\/} KL-divergence is desirable (e.g., by increasing its weight in a VAE's objective function, which is known as the $\beta$-VAE), as it imposes a stricter information bottleneck that forces the learned latent factors to be more independent (i.e., non-redundant), leading to a better disentangling. 

\begin{figure}[!tb]
\begin{center}
\centerline{\includegraphics[width=0.98\columnwidth]{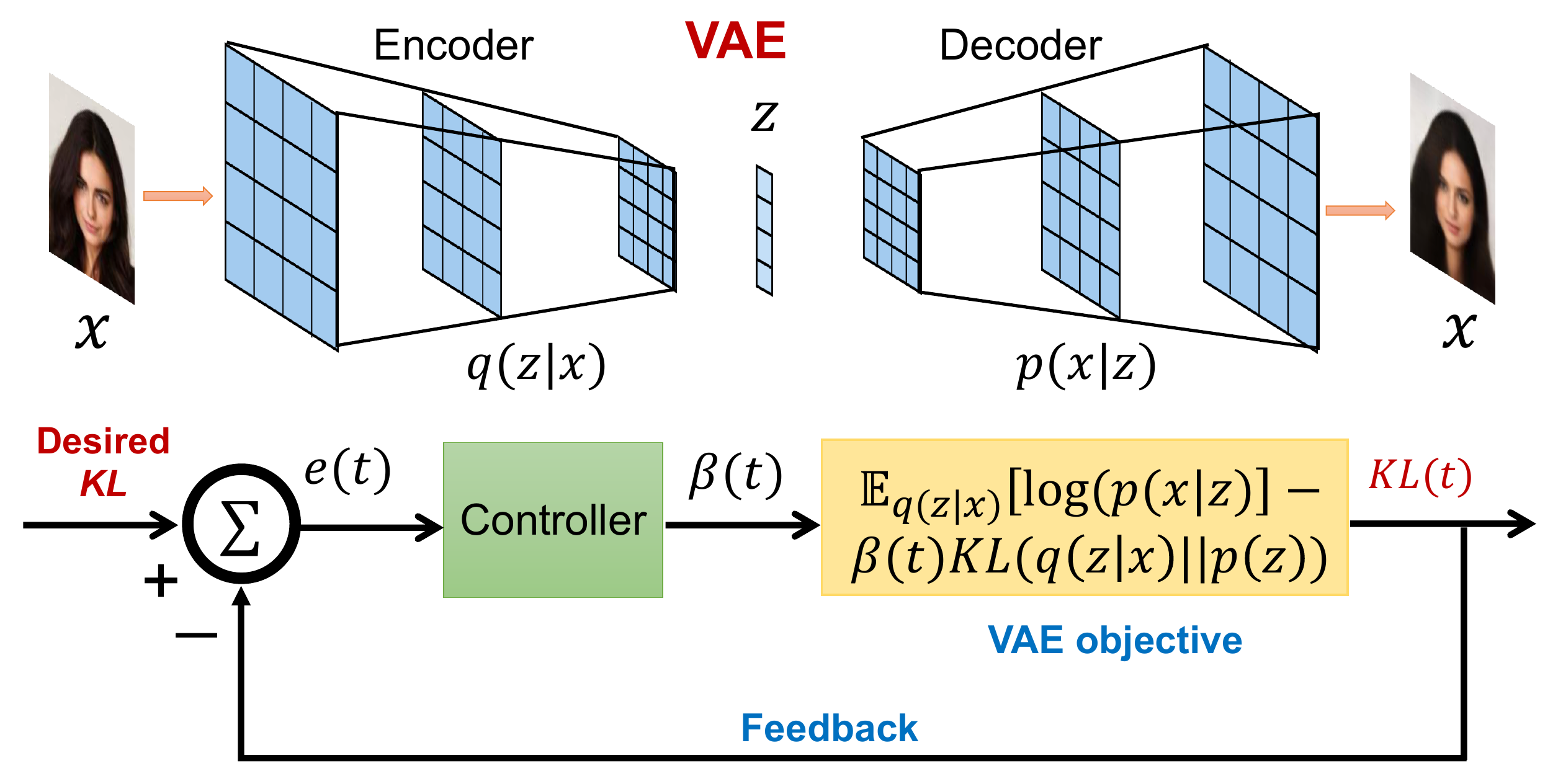}}
\caption{Framework of ControlVAE. It combines a controller with the basic VAE framework to stabilize the KL divergence to a specified value via automatically tuning the weight $\beta(t)$ in the objective.}
\label{fig:controlVAE}
\end{center}
\end{figure}

The above examples suggest that a useful extension of VAEs is one that allows users to explicitly control the KL-divergence term in the objective. Accordingly, much theoretical analysis was done to justify putting more emphasis on one of the two terms in the VAE's objective function, rather than treating them equally~\cite{rezende2018taming,klushyn2019learning,alemi2017fixing,zhao2019infovae,burgess2018understanding}. In this paper, we develop a systematic method \textit{method to control the KL-divergence}. Previous solutions mainly assign a fixed or learnable weight for the KL term~\cite{higgins2017beta, burgess2018understanding,alemi2017fixing,dai2019diagnosing,asperti2020balancing} to manipulate the value of KL-divergence. However, they cannot accurately control the value of the KL-divergence or achieve a good trade-off with reconstruction error. To address this issue, we propose a novel controllable variational autoencoder, ControlVAE, that leverages automatic control to control the trade-off between reconstruction accuracy and KL-divergence as shown in Fig.~\ref{fig:controlVAE}. Specifically, a non-linear PI controller is designed that stabilizes the value of KL-divergence via dynamically tuning the weight of the KL term during training. The reason why we adopt PI control algorithm instead of machine learning methods, such as Bayesian optimization, is that our method is an on-line dynamic tuning approach that only needs one-round training, while Bayesian optimization is an off-line profiling method that requires to train model multiple rounds with different weights.


This paper is an extension of work originally presented in ICML2020~\cite{shao2020controlvae}. This work is different from the prior paper in several respects. First, in order to determine its set point of KL-divergence that improves ELBO over regular VAE, we offer an analytic proof of this result. Then we further verify these analytical results empirically via conducting a new set of experiments on the task of image generation. In addition, we present the connection between ControlVAE and the existing VAE models as well as the constrained optimization of the KL term using Lagrange multiplier. In order to further improve its disentanglement ability, we explore a new variation of ControlVAE, called Control-FactorVAE, to improve the learning of disentangled representations. Since the prior metric, mutual information gap (MIG), is merely used to measure the overall score for disentanglement, we adopt another new robust MIG (RMIG)~\cite{do2020theory} to measure the disentanglment score of each generating factor. Finally, we conduct ablation studies to explore the effect of hyper-parameters on the performance of ControlVAE.

We apply our proposed methods to three different tasks: language modeling, disentangling, and image generation. Evaluation results on benchmark datasets demonstrate that ControlVAE is able to achieve an {\em adjustable trade-off\/} between reconstruction error and KL-divergence. It can significantly reduce the reconstruction error while achieving comparable disentanglement. We also show that ControlVAE is able to improve the ELBO and reconstruction quality on the task of image generation. For language modeling, it completely avoids the posterior collapse (KL vanishing) and improves the diversity of generated data.

%% file: preliminary.tex
\section{Preliminaries}
\label{sec:preliminary}
The objective function of VAEs consists of two terms: log-likelihood and KL-divergence. The first term tries to reconstruct the input data, while KL-divergence has the desirable effect of keeping the representation of input data sufficiently diverse. In particular, KL-divergence can affect both the reconstruction quality and diversity of generated data. If the KL-divergence is too high, it would affect the accuracy of generated samples. If it is too low, output diversity is reduced, which may be a problem in some applications such as language modeling~\cite{bowman2015generating} (where it is known as the KL-vanishing problem). 

To mitigate KL vanishing, one promising way is to add an extra hyperparameter $\beta (0 \leq \beta \leq 1)$ in the VAE objective function to control the KL-divergence via increasing $\beta$ from $0$ until to $1$ with sigmoid function or cyclic function~\cite{liu2019cyclical}. These methods, however, blindly change $\beta$ without sampling the actual KL-divergence during model training. Using a similar methodology, researchers recently developed a new $\beta$-VAE ($\beta > 1$)~\cite{higgins2017beta,burgess2018understanding} to learn the disentangled representations by controlling the value of KL-divergence. However, $\beta$-VAE suffers from high reconstruction errors~\cite{kim2018disentangling}, because it adds a very large $\beta$ in the VAE objective so that the model tends to focus disproportionately on optimizing the KL term. In addition, its hyperparameter is fixed during model training, missing the chance of balancing the reconstruction error and KL-divergence.

The core technical challenge responsible for the above application problems lies in the difficulty to tune the weight of the KL-divergence term during model training. Inspired by control systems, we fix this problem using feedback control. Our controllable variational autoencoder is illustrated in Fig.~\ref{fig:controlVAE}. It samples the output KL-divergence at each training step $t$, and feeds it into an algorithm that tunes the hyperparameter, $\beta(t)$, accordingly, aiming to stabilize KL-divergence at a desired value, called \textit{set point}. We further design a non-linear PI controller, a variant of the PID control algorithm~\cite{aastrom2006advanced}, to tune the hyperparameter $\beta(t)$. The advantage of PID control algorithm over Bayesian optimization is that it is a on-line dynamic tuning method that only needs one-round training, which has very low computational complexity. Next, we introduce the background of PID algorithm in detail.


PID control is the basic and most prevalent form of feedback control in a large variety of industrial~\cite{aastrom2006advanced} and software performance control~\cite{hellerstein2004feedback} applications. The general model of PID controller is defined by
\begin{equation}\label{eq:pid}
\beta(t) = K_p e(t) + K_i \int_0^t e(\tau)d\tau + K_d \frac{de(t)}{dt},
\end{equation}
\noindent
where $\beta(t)$ is the output of the controller; $e(t)$ is the error between the actual value and the desired value at time $t$; $K_p, K_i$ and $K_d$ denote the coefficients for the P term, I term and D term, respectively.

The basic idea of the PID algorithm is to calculate an error, $e(t)$, between a set point (in this case, the desired KL-divergence) and the current value of the controlled variable (in this case, the actual KL-divergence), then apply a correction in a direction that reduces that error. The correction is applied to some intermediate directly accessible variable (in our case, $\beta(t)$) that influences the value of the variable we ultimately want to control (KL-divergence). In general, the correction computed by the controller is the weighted sum of three terms; one changes with error (P), one changes with the integral of error (I), and one changes with the derivative of error (D). In a nonlinear controller, the changes can be described by {\em nonlinear\/} functions. Note that, since derivatives essentially compute the slope of a signal, when the signal is noisy, the slope often responds more to variations induced by noise. Hence, following established best practices in control of noisy systems, we do not use the derivative (D) term in our specific controller. 
Next, we introduce VAEs and our objective in more detail.

\subsection{The Variational Autoencoder (VAE)}
\label{sec:vae-intro}
The VAE \cite{kingma2013auto,rezende2014stochastic} has been one of the most popular types of generative models. It assumes a latent variable $\mathbf{z}$ with a prior $p(\mathbf{z})$, and a conditional distribution $p_{\theta}(\mathbf{x}|\mathbf{z})$, to model the observed variable $\mathbf{x}$. The generative model, denoted by $p_{\theta}(\mathbf{x})$, can be expressed as $p_{\theta}(\mathbf{x})=\int_{\mathcal{Z}} p_{\theta}(\mathbf{x} | \mathbf{z}) p(\mathbf{z})\mathrm{d} \mathbf{z}$. However, direct computation of this integral is often intractable, hence variational inference is used to derive a lower bound on $\log p_{\theta}(\mathbf{x})$. This leads to the evidence lower bound (ELBO):
\begin{equation}\label{eq:vae}
\begin{split}
& \log p_\theta (\mathbf{x})  \geq \log p_{\theta}(\mathbf{x})- KL(q_{\phi}(\mathbf{z} | \mathbf{x}) \| p_{\theta}(\mathbf{z} | \mathbf{x})) \\
& =  \mathbb{E}_{q_\phi(\mathbf{z|x)}} [\log p_\theta(\mathbf{x|z})] - KL(q_\phi(\mathbf{z|x})||p(\mathbf{z})) 
\\ & = \mathcal{L}_{recon}+\mathcal{L}_{KL} \triangleq \mathcal{L}_{vae},
\end{split}
\end{equation}
where $p_\theta (\mathbf{x}|\mathbf{z})$ is a probabilistic \textit{decoder} parameterized by a neural network to generate data $\mathbf{x}$ given the latent variable $\mathbf{z}$, and the posterior distribution of latent variable $\mathbf{z}$ given data $\mathbf{x}$ is approximated by the variational posterior, $q_{\phi} (\mathbf{z}|\mathbf{x})$, which is parameterized by an \textit{encoder} network. The VAE is trained by maximizing $\mathcal{L}_{vae}$, which consists of a reconstruction term and a KL term, over the training data.

However, plain VAEs cannot balance reconstruction error and KL-divergence, making them unsuitable to be applied to some specific tasks. For example, VAEs often suffer from KL vanishing in language modeling~\cite{bowman2015generating,liu2019cyclical}, meaning that the KL-divergence becomes nearly zero during optimization.
%

\subsection{$\beta$-VAE}
\label{sec:beta-vae}
The $\beta$-VAE~\cite{higgins2017beta,chen2018isolating} is an extension to the basic VAE framework, often used as an unsupervised method for learning a disentangled representation of the data generative factors. A disentangled representation, according to the literature~\cite{bengio2013representation}, is defined as one where single latent units are sensitive to changes in single generative factors, while being relatively invariant to changes in other factors.
Compared to the original VAE, $\beta$-VAE adds an extra hyperparameter $\beta (\beta >1 )$ as a weight of KL-divergence in the original VAE objective~\eqref{eq:vae}. It can be expressed by
\begin{equation}\label{eq:beta-vae}
\mathcal{L}_{\beta} = \mathbb{E}_{q_\phi(\mathbf{z|x)}} [\log p_\theta(\mathbf{x|z})] - \beta KL(q_\phi(\mathbf{z|x})||p(\mathbf{z})).
\end{equation}
In order to discover more disentangled factors, researchers further put a constraint on total information capacity, $C$, to control the capacity of the information bottleneck (KL-divergence)~\cite{burgess2018understanding}. Then Lagrangian method is adopted to solve the following optimization problem.
\begin{equation}\label{eq:improved-vae}\small
\mathcal{L}_{\beta} = \mathbb{E}_{q_\phi(\mathbf{z|x)}} [\log p_\theta(\mathbf{x|z})] - \beta | KL(q_\phi(\mathbf{z|x})||p(\mathbf{z})) -C |,
\end{equation}
where $\beta$ is a large hyperparameter (e.g., 100).

However, one drawback of $\beta$-VAE is that it obtains good disentangling at the cost of reconstruction quality. When the weight $\beta$ is large, the optimization algorithm tends to optimize the second term in \eqref{eq:improved-vae}, leading to a high reconstruction error.

\subsection{FactorVAE}
\label{sec:factorvae}
FactorVAE augments the VAE objective with an extra term that directly encourages independence in the latent code distribution. The objective of FactorVAE is given by
\begin{equation}
\begin{split}
\mathcal{L}_{f} = &\mathbb{E}_{q_\phi(\mathbf{z|x)}} [\log p_\theta(\mathbf{x|z})] - KL (q_\phi(\mathbf{z|x})||p(\mathbf{z})) \\
&- \gamma  KL (q(\mathbf{z})||\bar{q}(\mathbf{z})),
\end{split}
\end{equation}
where $q(\mathbf{z})$ is the distribution of representation for the entire input data, and $\bar{q}(\mathbf{z}) = \prod_{j=1}^d q(z_j)$ is defined to be the product of the marginal distribution of each latent variable $z_j$. The third term is called Total Correlation (TC)~\cite{watanabe1960information}, which measures the independence among different random variables. Note that $KL (q(\mathbf{z})||\bar{q}(\mathbf{z}) = 0$ if and only if each $z_j$ is independent under $q(\mathbf{z})$. In addition, $\gamma$ is a hyperparameter that represents the weight on the penalty of total correlation term. Since FactorVAE directly penalizes the total correlation term without constraining the mutual information between $\mathbf{x}$ and $\mathbf{z}$, it has better reconstruction quality than $\beta$-VAE. However, as the weight $\gamma$ is fixed, we still cannot explicitly control the value of TC term to ensure it is small enough after training.

The above background suggests that a common challenge in applying VAEs (and their extensions) lies in appropriate weight allocation among the reconstruction accuracy and KL-divergence in the VAEs objective function. As mentioned earlier, we solve this using a nonlinear PI controller that manipulates the value of the non-negative hyperparameter, $\beta(t)$. This algorithm is described next.

%% file: model.tex
\section{The ControlVAE Algorithm}
\label{sec:model}
During model training, we sample the output KL-divergence, which we denote by $\hat{v}_{kl}(t)$, at training step $t$. The sampled KL-divergence is then compared to the set point, ${v}_{kl}$, and the difference, $e(t) = {v}_{kl} - \hat{v}_{kl}(t)$, then used as the feedback to a controller to calculate the hyperparameter $\beta(t)$. ControlVAE can be expressed by the following variational lower bound:
\begin{equation}\label{eq:controlVAE-object}
\mathcal{L} = \mathbb{E}_{q_\phi(\mathbf{z}|\mathbf{x})} [\log p_\theta(\mathbf{x|z})] - \beta(t) KL(q_\phi(\mathbf{z}|\mathbf{x})||p(\mathbf{z})).
\end{equation}

When KL-divergence drops below the set point, the controller counteracts this change by reducing the hyperparameter $\beta(t)$ (to reduce penalty for KL-divergence in the objective function~\eqref{eq:controlVAE-object}). The reduced weight, $\beta(t)$, allows KL-divergence to grow, thus approaching the set point again. Conversely, when KL-divergence grows above the set point, the controller increases $\beta(t)$ (up to a certain value), thereby increasing the penalty for KL-divergence and forcing it to decrease. This effect is achieved by computing $\beta(t)$ using Equation~(\ref{eq:vae-lang}), below, which is an instance of nonlinear PI control: \begin{equation}\label{eq:vae-lang}
\vspace{-0.02in}
\beta(t) = \frac{K_p}{1+\exp(e(t))} - K_i \sum_{j=0}^t e(j) + \beta_{min},
\end{equation}
where $K_p$ and $K_i$ are the constants. The first term (on the right hand side) ranges between $0$ and $K_p$ thanks to the exponential function $exp(.)$. Note that when error is large and positive (KL-diverge is below set point), the first term approaches 0, leading to a lower $\beta(t)$ that encourages KL-divergence to grow. Conversely, when error is large and negative (KL-divergence above set point), the first term approaches its maximum (which is $K_p$), leading to a higher $\beta(t)$ that encourages KL-divergence to shrink.

\begin{figure}[!t]
\begin{center}
\centerline{\includegraphics[width=0.99\columnwidth]{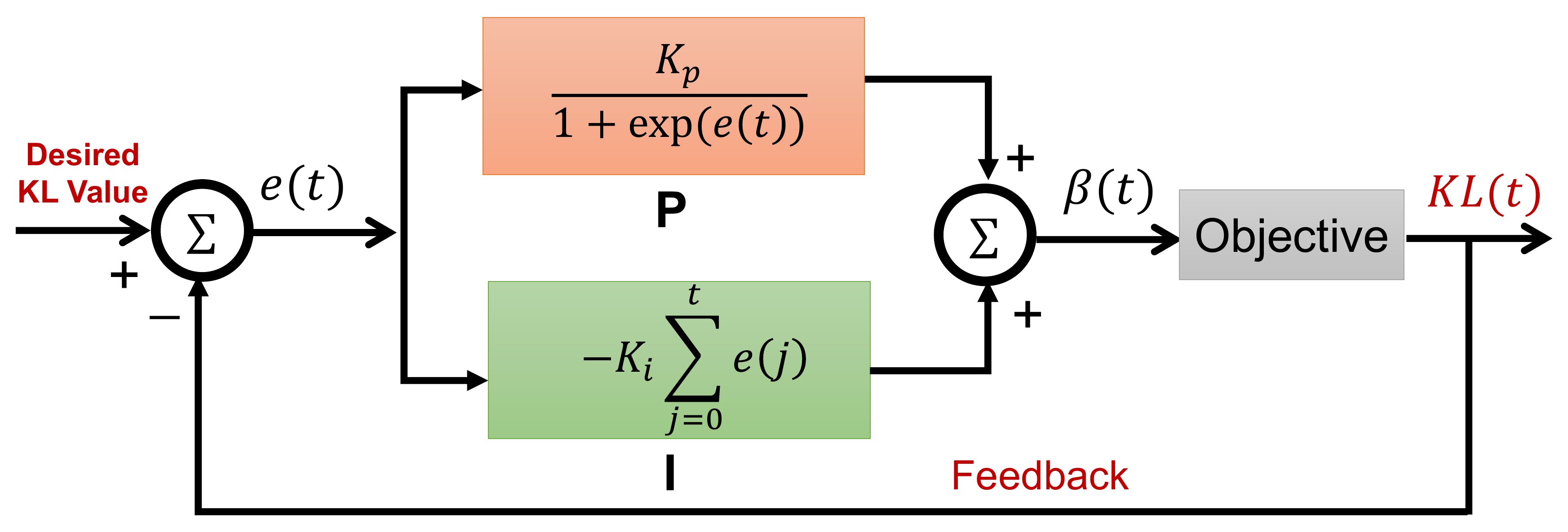}}
\caption{The block diagram of PI controller. It uses the output KL-divergence at training step $t$ as the feedback to the PI algorithm to compute $\beta(t)$.}
\label{fig:pid}
\end{center}
\end{figure}

The second term of the controller sums (integrates) past errors with a sampling period $T$ (one training step in this paper). This creates a progressively stronger correction (until the sign of the error changes). The negative sign ensures that while errors remain positive (i.e., when KL-divergence is below set point), this term continues to decrease, whereas while errors remain negative (i.e., when KL-divergence is above set point), this term continues to increase. In both cases, the change forces $\beta(t)$ in a direction that helps KL-divergence approach the set point. In particular, note that when the error becomes zero, the second term (and thus the entire right hand side) stops changing, allowing controller output, $\beta(t)$, to stay at the same value that hopefully caused the zero error in the first place. This allows the controller to ``lock in" the value of $\beta(t)$ that meets the KL-divergence set point.
Finally, $\beta_{min}$ is an application-specific constant. It effectively shifts the range within which $\beta(t)$ is allowed to vary. This PI controller is illustrated in Fig.~\ref{fig:pid}.

\subsection{The Controllable FactorVAE Algorithm}
\label{sec:control-factorVAE}
For disentangled representation learning, the total correlation (TC) term of FactorVAE may collapse to 0, leading to bad disentanglement. To deal with this issue, we propose a novel controllable FactorVAE, Control-FactorVAE, to stabilize the TC to a small value based on the actual value of TC during model training. Similar to ControlVAE, the objective function of Control-FactorVAE is expressed as
\begin{equation}\label{eq:control-factor}
\begin{split}
\mathcal{L}_{f} & =  \mathbb{E}_{q_\phi(\mathbf{z|x)}} [\log p_\theta(\mathbf{x|z})] - KL (q_\phi(\mathbf{z|x})||p(\mathbf{z})) \\
&- \beta(t)  KL (q(\mathbf{z})||\bar{q}(\mathbf{z})),
\end{split}
\end{equation}
where $\beta(t)$ is the output of the above designed PI controller using the output TC as feedback.

\subsection{PI Parameter Tuning for ControlVAE}
\label{sec:pi_tuning}
One challenge of applying the PI control algorithm lies how to tune its parameters, $K_p$ and $K_i$ effectively. While optimal tuning of nonlinear controllers is non-trivial, in this paper we follow a very simple rule: tune these constants to ensure that reactions to errors are sufficiently smooth to allow gradual convergence.
Let us first consider the coefficient $K_p$. Observe that the maximum (positive) error occurs when actual KL-divergence is close to zero. In this case, if $v_{kl}$ is the set point on KL-divergence, then the error, $e(t)$, is approximated by $e(t) \approx v_{kl} - 0 = v_{kl}$. When KL-divergence is too small, the VAE does not learn useful information from input data~\cite{liu2019cyclical}. We need to assign $\beta(t)$ a very small non-negative value, so that KL-divergence is encouraged to grow (when the resulting objective function is optimized). In other words, temporarily ignoring other terms in Equation~\eqref{eq:vae-lang}, the contribution of the first term alone should be sufficiently small:
\begin{equation}\label{eq:Kp}
\frac{K_p}{1+\exp( v_{kl} )} \leq \epsilon,
\end{equation}
where $\epsilon$ is a small constant (e.g., $10^{-3}$ in our implementation). The above~\eqref{eq:Kp} can also be rewritten as $K_p \leq (1+\exp(v_{kl}))\epsilon$. Empirically, we find that $K_p=0.01$ leads to good performance and satisfies the above constraint.

Conversely, when the actual KL-divergence is much larger than the desired value $v_{kl}$, the error $e(t)$ becomes a large negative value. As a result, the first term in~\eqref{eq:vae-lang} becomes close to a constant, $K_p$. If the resulting larger value of $\beta(t)$ is not enough to cause KL-divergence to shrink, one needs to gradually continue to increase $\beta(t)$. This is the job of second term. 
The negative sign in front of that term ensures that when negative errors continue to accumulate, the positive output $\beta(t)$ continues to increase. Since it takes lots of steps to train deep VAE models,
the increase per step should be very small, favoring smaller values of $K_i$. 
Empirically we found that a value $K_i$ between $10^{-3}$ and $10^{-4}$ stabilizes the training. Note that, $K_i$ should not be too small either, because it would then unnecessarily slow down the convergence.

\subsection{Set Point Guidelines for ControlVAE}
\label{sec:bound}
The choice of desired value of KL-divergence (set point) is largely application specific. In general, when $ \beta_{min} \leq \beta(t) \leq \beta_{max}$, the upper bound of expected KL-divergence is the value of KL-divergence as ControlVAE converges when $\beta(t) = \beta_{min}$, denoted by $V_{max}$. Similarly, its lower bound, $V_{min}$, can be defined as the KL-divergence produced by ControlVAE when $\beta(t) = \beta_{max}$. For feedback control to be most effective (i.e., not run against the above limits), the KL-divergence set point should vary in the range of $[V_{min}, V_{max}]$. Since ControlVAE is an end-to-end learning model, users can customize the desired value of KL-divergence (using KL-divergence of the original VAE as a reference) to meet their demand with respect to different applications. For instance, if some users prefer to improve the diversity of text generation and image generation, they can slightly increase the KL-divergence produced by the original VAE. Otherwise they can reduce the KL-divergence if they want to improve the generation accuracy.

In this paper, we provide a simplified theoretical analysis about how to choose the set point if our goal is to improve the ELBO or the reconstruction accuracy over the basic VAE. Let $d$ ($d>0$) denote the optimal range of KL-divergence between ControlVAE and the basic VAE, and $KL_{vae}$ denote the value of KL-divergence as the basic VAE converges. Then the KL-divergence of ControlVAE can be denoted by
\begin{equation}
KL_{cvae} = KL_{vae} + d.
\end{equation}
\noindent
Therefore, the ELBO of the ControlVAE can be written as
\begin{equation}
\mathcal{L}_{cvae} =  \mathbb{E}_{q_\phi(\mathbf{z}|\mathbf{x})} [\log p_\theta(\mathbf{x|z})] - (KL_{vae} + d)
\end{equation}

In order to achieve a higher ELBO for ControlVAE, we want
\begin{equation}\label{eq:elbo_compare}
\begin{split}
 & \mathcal{L}_{cvae}  - \mathcal{L}_{vae}  =  \\
 &\mathbb{E}_{q_\phi(\mathbf{z}|\mathbf{x})} [\log p_\theta(\mathbf{x|z})] - \mathbb{E}_{q_{\phi'}(\mathbf{z_0|x)}} [\log p_{\theta'}(\mathbf{x|z}_0)] -d \geq 0,
 \end{split}
\end{equation}
where $\mathbf{z}$ and $\mathbf{z}_0$ denote the latent variable of ControlVAE and the original VAE, respectively. Since the KL term mainly affects the encoder parameters, to simplify analysis, we may assume that $\theta \approx \theta'$. 

In order to obtain the range $d$, our next step is to bound the difference of reconstruction accuracy between ControlVAE and the basic VAE, $\mathbb{E}_{q_\phi(\mathbf{z}|\mathbf{x})} [\log p_\theta(\mathbf{x|z})] - \mathbb{E}_{q_\phi(\mathbf{z_0|x)}} [\log p_\theta(\mathbf{x|z}_0)]$. Assuming the decoder to be Lipschitz continuous~\cite{virmaux2018lipschitz}, we have the following theorem.
\begin{theorem}\label{the:z_eq}
Given the KL-divergence, $KL_{vae}$, of the original VAE as it converges, we have
\begin{equation}\label{eq:z_range}\small
\mathbb{E}_{q_\phi(\mathbf{z}|\mathbf{x})} [\log p_\theta(\mathbf{x|z})] - \mathbb{E}_{q_\phi(\mathbf{z_0|x)}} [\log p_\theta(\mathbf{x|z}_0)]  \leq   \sqrt{ 4(2KL_{vae}+d)}.
\end{equation}
\end{theorem}
\noindent
\textit{Proof:} please see it in Appendix~\ref{app:theorem1}.

\noindent
Based on the above Theorem~\ref{the:z_eq} and Eq.~\eqref{eq:elbo_compare}, we have
\begin{equation}\label{eq:lipt_eta}
0 \leq d \leq \sqrt{ 4(2KL_{vae}+d)},
\end{equation}
From the above equation, we can derive the range $d$:
\begin{equation}
0  \leq d \leq 2 + 2\sqrt{2KL_{vae} +1 }.
\end{equation}
Therefore, if we choose $d$ in the above range, we can improve the ELBO of ControlVAE.

\subsection{Summary of the PI Control Algorithm}
We summarize the proposed PI control algorithm in Algorithm~\ref{alg:pid}. Our PI algorithm updates the hyperparameter, $\beta(t)$, with the feedback from sampled KL-divergence at training step $t$. Line $6$ computes the error between the desired KL-divergence, $v_{kl}(t)$, and the sampled $\hat{v}_{kl}(t)$. Line $7$ to $9$ calculate the P term and I term for the PI algorithm, respectively. Note that, Line 10 and 11 is a popular constraint in PID/PI design, called anti-windup~\cite{azar2015design,peng1996anti}. It effectively disables the integral term of the controller when controller output gets out of range, not to exacerbate the out-of-range deviation. Line $13$ is the calculated hyperparameter $\beta(t)$ by PI algorithm in~\eqref{eq:vae-lang}. Finally, Line $14$ to $19$ aim to limit $\beta(t)$ to a certain range, $[\beta_{min},\beta_{max}]$.

\begin{algorithm}[!tb]
   \caption{PI algorithm.}
   \label{alg:pid}
\begin{algorithmic}[1]
   \STATE {\bfseries Input:} desired KL $v_{kl}$, coefficients $K_p$, $K_i$, max/min value $\beta_{max}$, $\beta_{min}$, iterations $N$
   \STATE {\bfseries Output:} hyperparameter $\beta(t)$ at training step $t$
   \STATE {\bfseries Initialization}: $I(0)=0$, $\beta(0)=0$
   \FOR{$t=1$ {\bfseries to} $N$}
   \STATE Sample KL-divergence, $\hat{v}_{kl}(t)$
   \STATE $e(t) \leftarrow v_{kl}-\hat{v}_{kl}(t)$
   \STATE $P(t) \leftarrow \frac{K_p}{1+\exp(e(t))}$
   \IF{ $ \beta_{min} \leq \beta(t-1) \leq \beta_{max}$ }
   		\STATE $I(t) \leftarrow I(t-1) - K_i e(t)$
   \ELSE
   \STATE $I(t) \leftarrow I(t-1)$ \quad // Anti-windup
	 \ENDIF
   \STATE $\beta(t) \leftarrow P(t)+I(t) + \beta_{min}$
   \IF{$\beta(t)> \beta_{max}$}
   \STATE $\beta(t) \leftarrow \beta_{max}$
 	\ENDIF
   \IF{$\beta(t)< \beta_{min}$}
   \STATE $\beta(t) \leftarrow \beta_{min}$
   \ENDIF
   \STATE \textbf{Return} $\beta(t)$
 \ENDFOR
\end{algorithmic}
\end{algorithm}


\subsection{Connection to Other Models}\label{sec:connect}
We are going to illustrate the connection between ControlVAE and some other existing VAE models.
\subsubsection{Connection to Lagrange Multiplier}
The goal of ControlVAE is to dynamically tune the weight on KL term to stabilize the KL-divergence to a desired value, $v_{kl}$. Inspired by literature~\cite{rezende2018taming}, we can also formulate it as a constrained optimization problem
\begin{equation}
\begin{aligned}
\max_{\phi, \theta} \quad &  \mathbb{E}_{q_\phi(\mathbf{z}|\mathbf{x})} [\log p_\theta(\mathbf{x|z})]\\
\textrm{s.t.} \quad & KL(q_\phi(\mathbf{z}|\mathbf{x})||p(\mathbf{z})) - v_{kl} = 0\\
\end{aligned}
\end{equation}
The Lagrangian of this optimization problem is
\begin{equation}
\mathcal{L}_m = \mathbb{E}_{q_\phi(\mathbf{z}|\mathbf{x})} [\log p_\theta(\mathbf{x|z})] - \lambda (KL(q_\phi(\mathbf{z}|\mathbf{x})||p(\mathbf{z})) - v_{kl}),
\end{equation}
where $\lambda$ is the Lagrange multiplier.

\noindent
The above problem could be solved by classic gradient descent-ascent. In particular, the gradient with respect to the Lagrange multiplier $\lambda$ at the $i$-th iteration is given by
\begin{equation}
d\lambda {(i)} = \frac{\partial \mathcal{L}_m }{\partial \lambda} = v_{kl} - KL(q_\phi(\mathbf{z}|\mathbf{x})||p(\mathbf{z})) = e(i),
\end{equation}
where $e(i)$ is the error between the desired KL-divergence and the actual one, as defined in Section~\ref{sec:model}.
\noindent
At training step $i$, we can update $\lambda$ as follows:
\begin{equation}
\begin{split}
\lambda {(i)} & = \lambda {(i-1)} - \alpha d\lambda {(i)} \\
& = \lambda {(i-1)} - \alpha e(i)
\end{split}
\end{equation}
where $\alpha$ is the learning rate.

\noindent
After training the above model with $t$ iterations, the hyperparameter, $\lambda$, can be expressed by
\begin{equation}\label{eq:lambda}
\lambda {(t)} = \lambda {(0)} -\alpha \sum_{i=0}^t  e(i).
\end{equation}
When $\lambda$ is initialized with 0, the above Eq.~\eqref{eq:lambda} becomes
\begin{equation}
\lambda {(t)} = -\alpha \sum_{i=0}^t  e(i).
\end{equation}
\noindent
It can be observed from the above formula that $\lambda(t)$ is the same as I term of the designed PI algorithm in Eq.\eqref{eq:vae-lang} when $\alpha=K_i$. Thus, the optimization problem using Lagrange multiplier can be seen as a special case of ControlVAE. We also conduct experiments to compare the performance of ControlVAE with Lagrange multiplier method in Section~\ref{sec:evaluate}.

\subsubsection{Connection to VAE and $\beta$-VAE}
For the basic VAE, we have $\beta = 1$ on the KL term in the VAE objective. After model training, the KL-divergence of the basic VAE converges to a value, $KL_{vae}$.  When we set the target KL-divergence, $v_{kl}$, to $KL_{vae}$, ControlVAE becomes the basic VAE as the weight $\beta(t)$ converges to $1$ at the end of model training. For the $\beta$-VAE, it assigns a large and fixed weight to the KL term. As long as we fix the output of PI controller as a large value, ControlVAE then becomes $\beta$-VAE.

\subsection{Applications of ControlVAE}
As a preliminary demonstration of the general applicability of the above approach and as an illustration of its customizability, we apply ControlVAE to three different applications stated below.
\begin{itemize}
\item \textbf{Disentangling}: We then apply the ControlVAE model to achieve a better trade-off between reconstruction quality and disentangling. As mentioned in Section~\ref{sec:beta-vae}, $\beta$-VAE ($\beta>1$) assigns a large hyperparameter to the objective function to control the KL-divergence (information bottleneck), which, however, leads to a large reconstruction error. To mitigate this issue, we adopt ControlVAE to automatically adjust the hyperparameter $\beta(t)$ based on the output KL-divergence during model training. Using the similar methodology in~\cite{burgess2018understanding}, we train a single model by gradually increasing KL-divergence from $0.5$ to a desired value $C$ with a step function $\alpha$ for every $K$ training steps. Since $\beta(t)>1$, we set $\beta_{min}$ to $1$ for the PI algorithm in~\eqref{eq:vae-lang}. Following the PI tuning method above, the coefficients $K_p$ and $K_i$ are set to $0.01$ and $0.001$, respectively.

\item \textbf{Image generation}: In this paper, we try to leverage ControlVAE to manipulate (slightly increase) the value of KL-divergence to improve the ELBO and reconstruction quality over the basic VAE for image generation. Different from the original VAE ($\beta(t)=1$), we extend the range of the hyperparameter, $\beta(t)$, from $0$ to $1$ in our controlVAE model. Given a desired KL-divergence, controlVAE can automatically tune $\beta(t)$ within that range. For this task, we use the same PI control algorithm and hyperparameters as the above language modeling.

\item \textbf{Language modeling}: We first apply ControlVAE to solve the KL vanishing problem meanwhile improve the diversity of generated data. As mentioned in Section~\ref{sec:vae-intro}, the VAE models often suffer from KL vanishing in language modeling. The existing methods cannot completely solve the KL vanishing problem or explicitly manipulate the value of KL-divergence. In this paper, we adopt ControlVAE to control KL-divergence to a specified value to avoid KL vanishing using the output KL-divergence. Following PI tuning strategy in Section~\ref{sec:pi_tuning}, we set $K_p$, $K_i$ of the PI algorithm in~\eqref{eq:vae-lang} to $0.01$ and $0.0001$, respectively. In addition, $\beta_{min}$ is set to $0$ and the maximum value of $\beta(t)$ is limited to $1$.

\end{itemize}

\begin{figure*}[!t]
\centering     
\subfigure[Reconstruction loss]{\includegraphics[width=0.32\textwidth]{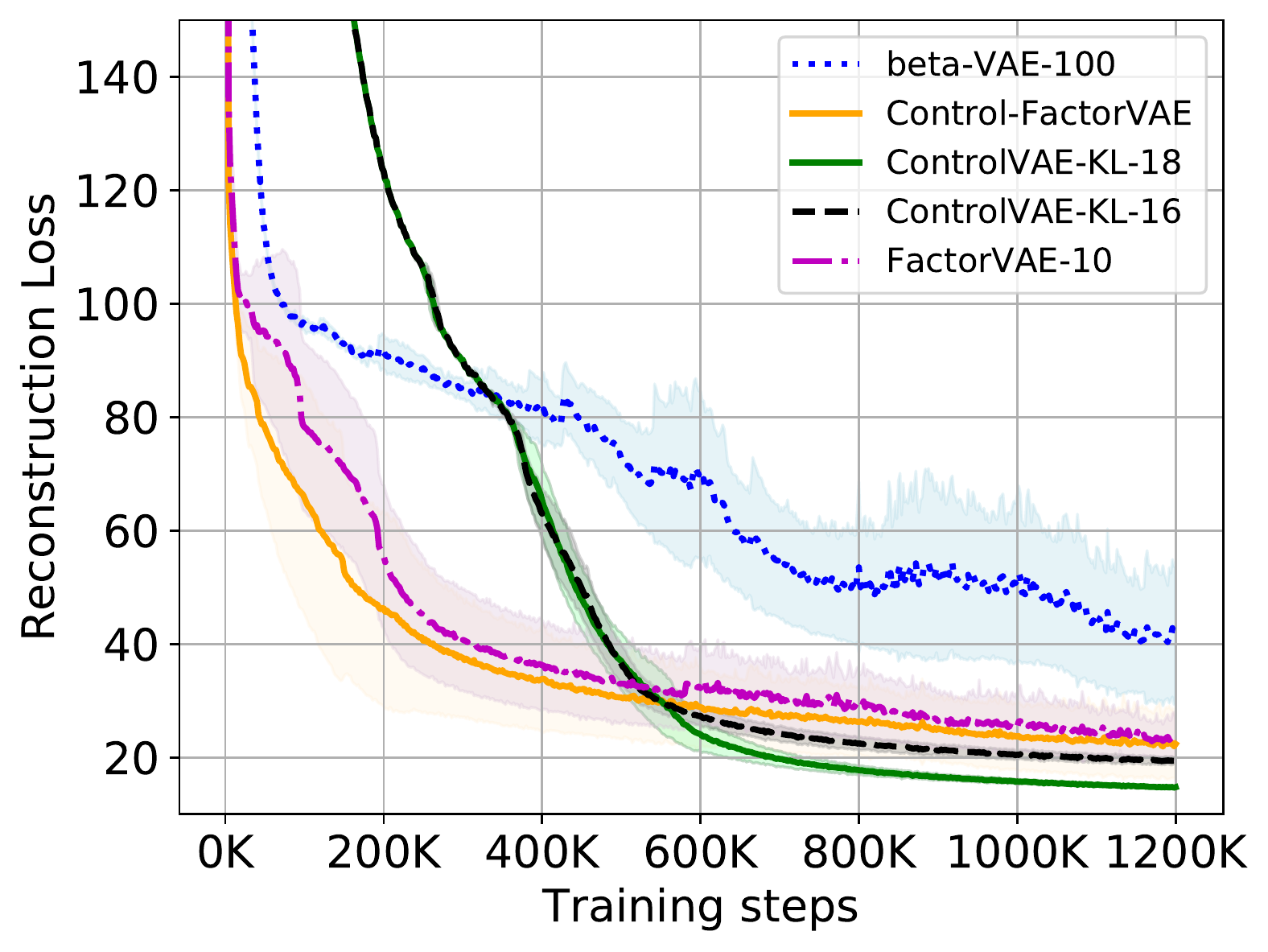}}
\subfigure[$\beta(t)$]{\includegraphics[width=0.32\textwidth]{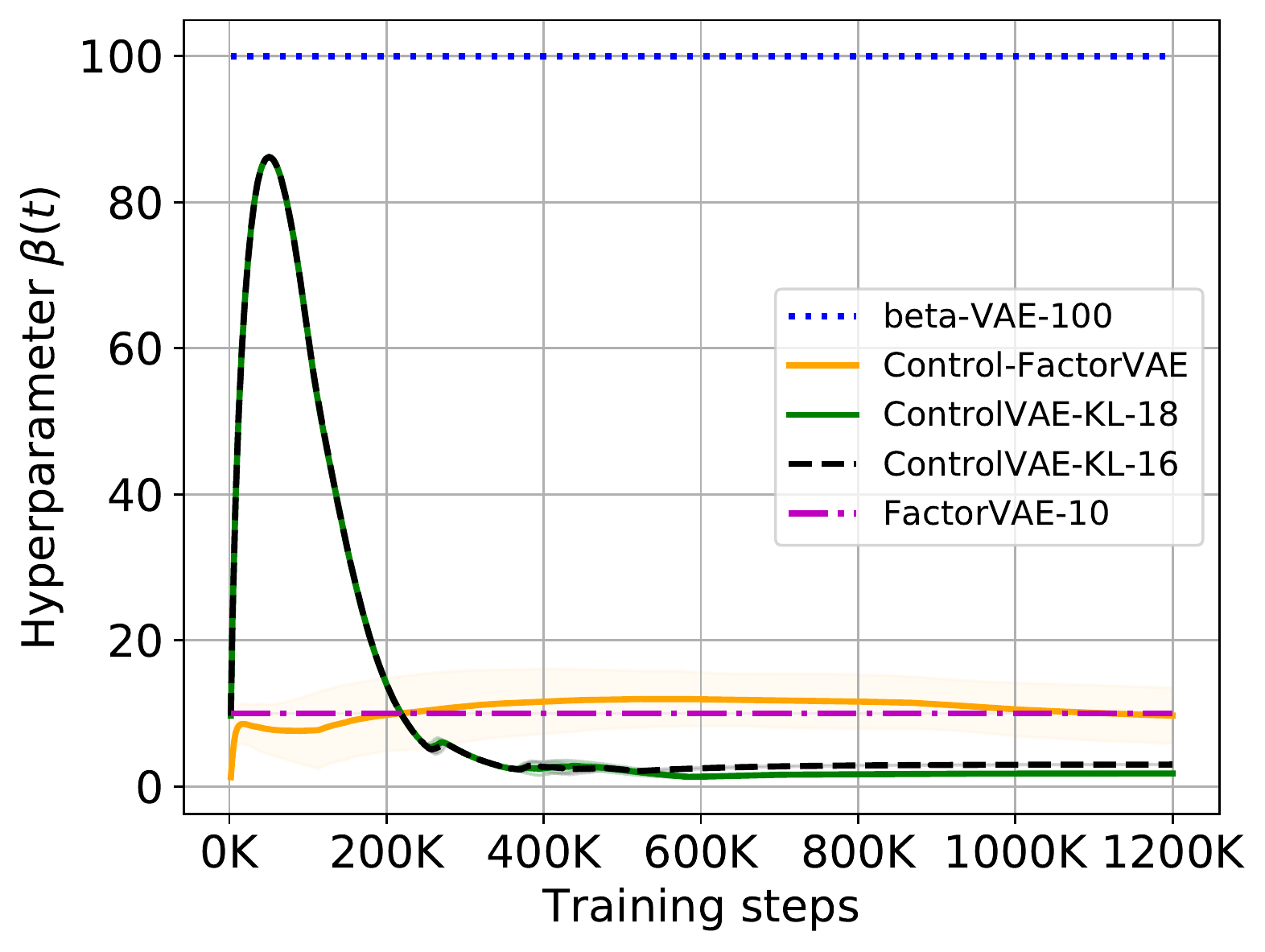}}
\subfigure[Disentangled factors]{\includegraphics[width=0.32\textwidth]{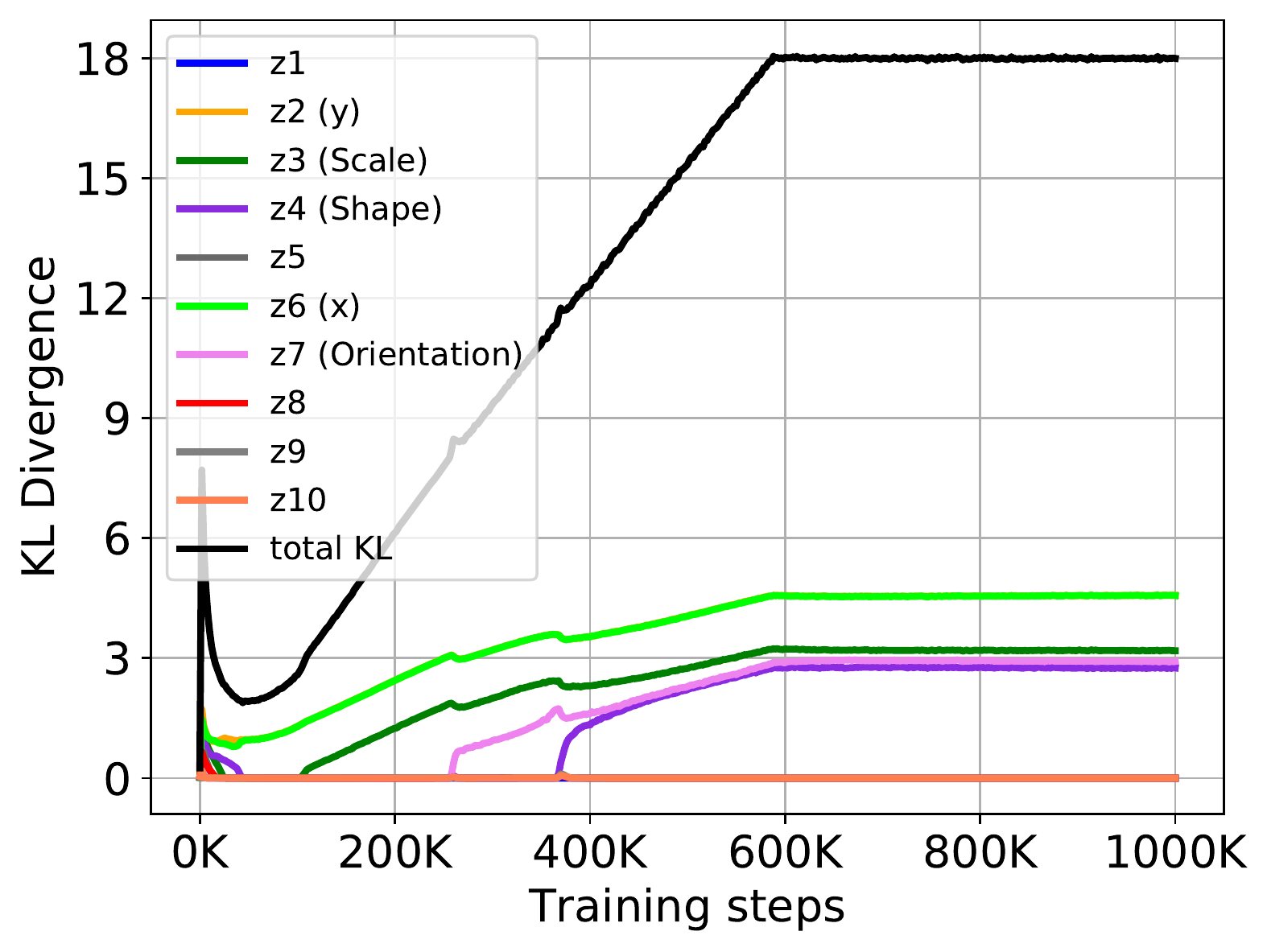}}
\caption{(a) (b) shows the comparison of reconstruction error and $\beta(t)$ using 2D Shapes data over $5$ random seeds. ControlVAE (KL=16, 18) and Control-FactorVAE have lower reconstruction errors and variance compared to the other methods. (c) shows an example about the disentangled factors in the latent variable as the total KL-divergence increases from $0.5$ to $18$ for ControlVAE (KL=18). Each curve with positive KL-divergence (except black one) represents one disentangled factor by ControlVAE.}\label{fig:disentangled}
\end{figure*}

%% file: Evaluation.tex
\section{Experiments}
\label{sec:evaluate}
We evaluate the performance of ControlVAE on benchmark datasets in the three different applications mentioned above. 


\subsection{Datasets}
The datasets used for our experiments are introduced below.
\begin{itemize}
\item Disentangling: 1)~\textbf{2D Shapes}~\cite{matthey2017dsprites}: it has $737,280$ binary $64 \times 64$ images of 2D shapes with five ground truth factors (number of values): shape(3), scale(6), orientation(40), x-position(32), y-position(32)~\cite{kim2018disentangling}.
\item Image generation: 1) \textbf{CelebA}(cropped version)~\cite{liu2015deep}: It has $202,599$ RGB $128 \times 128 \times 3$ images of celebrity faces. The data is split into $192,599$ and $10,000$ images for training and testing.
\item Language modeling: 1) \textbf{Penn Tree Bank (PTB)}~\cite{marcus1993building}: it consists of $42,068$ training sentences, $3,370$ validation sentences and $3,761$ testing sentences. 2) \textbf{Switchboard(SW)}~\cite{godfrey1997switchboard}: it has $2400$ two-sided telephone conversations with manually transcribed speech and alignment. The data is randomly split into $2316$, $60$ and $62$ dialog for training, validation and testing.
\end{itemize}

\subsection{Model Configurations}
The detailed model configurations and hyperparameter settings for each model is presented in Appendix~\ref{sec:configure}.

\subsection{Evaluation on Disentangled Representations}
First of all, we evaluate the performance of ControlVAE and Control-FactorVAE on the learning of disentangled representations using \textit{2D Shapes} data. We compare it with two baselines: FactorVAE~\cite{kim2018disentangling} and $\beta$-VAE~\cite{burgess2018understanding}.

Fig.~\ref{fig:disentangled} (a) and (b) shows the comparison of reconstruction error and the hyperparameter $\beta(t)$ (using $5$ random seeds) for different models. We can observe from Fig.~\ref{fig:disentangled} (a) that ControlVAE (KL=16,18) has lower reconstruction error and variance than the other methods. This is because our ControlVAE automatically adjusts the hyperparameter, $\beta(t)$, to stabilize the KL-divergence, while $\beta$-VAE and FactorVAE keep the hyperparameter unchanged during model training. In addition, the newly proposed Control-FactorVAE has slightly lower reconstruction error than the two baselines. Specifically, for ControlVAE (KL=18), the hyperparameter $\beta(t)$ is large in the beginning in order to obtain good disentangling, and then it gradually drops to around $1.8$ to improve reconstruction quality as the training converges, as shown in Fig.~\ref{fig:disentangled}(b). In contrast, $\beta$-VAE ($\beta=100$) and FactorVAE have a large and fixed weight on the KL-divergence in the objective so that its optimization algorithm tends to optimize the KL-divergence term, leading to a large reconstruction error. What is more, Fig.~\ref{fig:disentangled}(c) illustrates an example of KL-divergence per factor in the latent code as training progresses and the total information capacity (KL-divergence) increases from $0.5$ until to $18$. We can see that ControlVAE disentangles all the five generative factors, starting from positional latents ($x$ and $y$) to scale, followed by orientation and then shape.

Next, we use two disentanglement metrics, mutual information gap (MIG)~\cite{chen2018isolating} and robust MIG (RMIG) \cite{do2020theory}, to evaluate the disentanglement of different models. Table~\ref{tab:mig} illustrates the comparison of MIG score for different methods. It can be observed that ControlVAE (KL=16) has a comparable MIG but lower variance than FactorVAE and Control-FactorVAE. Here it is worth noting that FactorVAE and Control-FactorVAE add a Total Correlation (TC) term in the objective while ControlVAE does not. Besides, Control-FactorVAE (TC=0.3) and FactorVAE have comparable MIG scores, because they have the approximately equal weights after the models converge during training. We use MIG to measure average disentanglement score of four disentangled factors (scale, rotation, $x$ position and $y$ position), and we adopt RMIG, which is a robust version of MIG, to measure the score of each disentangled factor, as shown in Table~\ref{tab:rmig}. We can observe that Control-FactorVAE has the highest average RMIG score among them, because it can better disentangle the shape and scale factors. In addition, Control-FactorVAE, ControlVAE and FactorVAE have comparable RMIG score on average. For the performance of each single factor, ControlVAE (KL=16) has a better disentanglement than the other methods in terms of positional $x$ and $y$ while FactorVAE disentangles the orientation well.

\begin{table*}[!th]
\caption{Performance comparison of different methods using MIG score, averaged over 5 random seeds. The higher is better. ControlVAE (KL=16) has a comparable MIG score but lower variance than the FactorVAE with the default parameters.}
\label{tab:mig}
\begin{center}
\begin{small}
\begin{tabular}{llllll}
\toprule
Metric & ControlVAE (KL=16) & ControlVAE (KL=18) & Control-FactorVAE & $\beta$-VAE ($\beta=100$) & FactorVAE ($\gamma=10$) \\
\midrule
MIG & \textbf{ 0.5628 $\pm$ 0.0222}& 0.5432 $\pm$ 0.0281 &0.5620 $\pm$0.0348 & 0.5138 $\pm$ 0.0371 & 0.5625 $\pm$ 0.0443 \\
\bottomrule
\end{tabular}
\end{small}
\end{center}
\end{table*}

\begin{table*}[!th]
\caption{Comparison of RMIG for different methods averaged over 5 random seeds. The higher is better. Control-FactorVAE has a higher RMIG score than the other methods.}
\label{tab:rmig}
\begin{center}
\begin{small}
\begin{tabular}{lllllll}
\toprule
Metric & pos. $x$ & pos. $y$ & Shape & Scale & Orientation  & RMIG \\
\midrule
ControlVAE (KL=16)  & \textbf{0.7490} & \textbf{0.7413} & 0.0958 & 0.6288 & 0.1037  & 0.4637\\
ControlVAE (KL=18)  & 0.7229 & 0.7221 & 0.0701 & 0.6404 & 0.0582  &  0.4427\\
Control-FactorVAE & 0.7273 & 0.7442 & \textbf{0.1467} & \textbf{0.6996} & 0.0680  & \textbf{0.4772}\\
$\beta$-VAE ($\beta=100$) & 0.7112 & 0.7014  & 0.1325 & 0.5386 & 0.0819 & 0.4331\\
FactorVAE ($\gamma=10$) & 0.7482 & 0.7276 & 0.1383 & 0.6262 & \textbf{0.1412}  & 0.4763\\
\bottomrule
\end{tabular}
\end{small}
\end{center}
\end{table*}

\begin{figure*}[!th]
\begin{center}
\centerline{\includegraphics[width=0.94\textwidth]{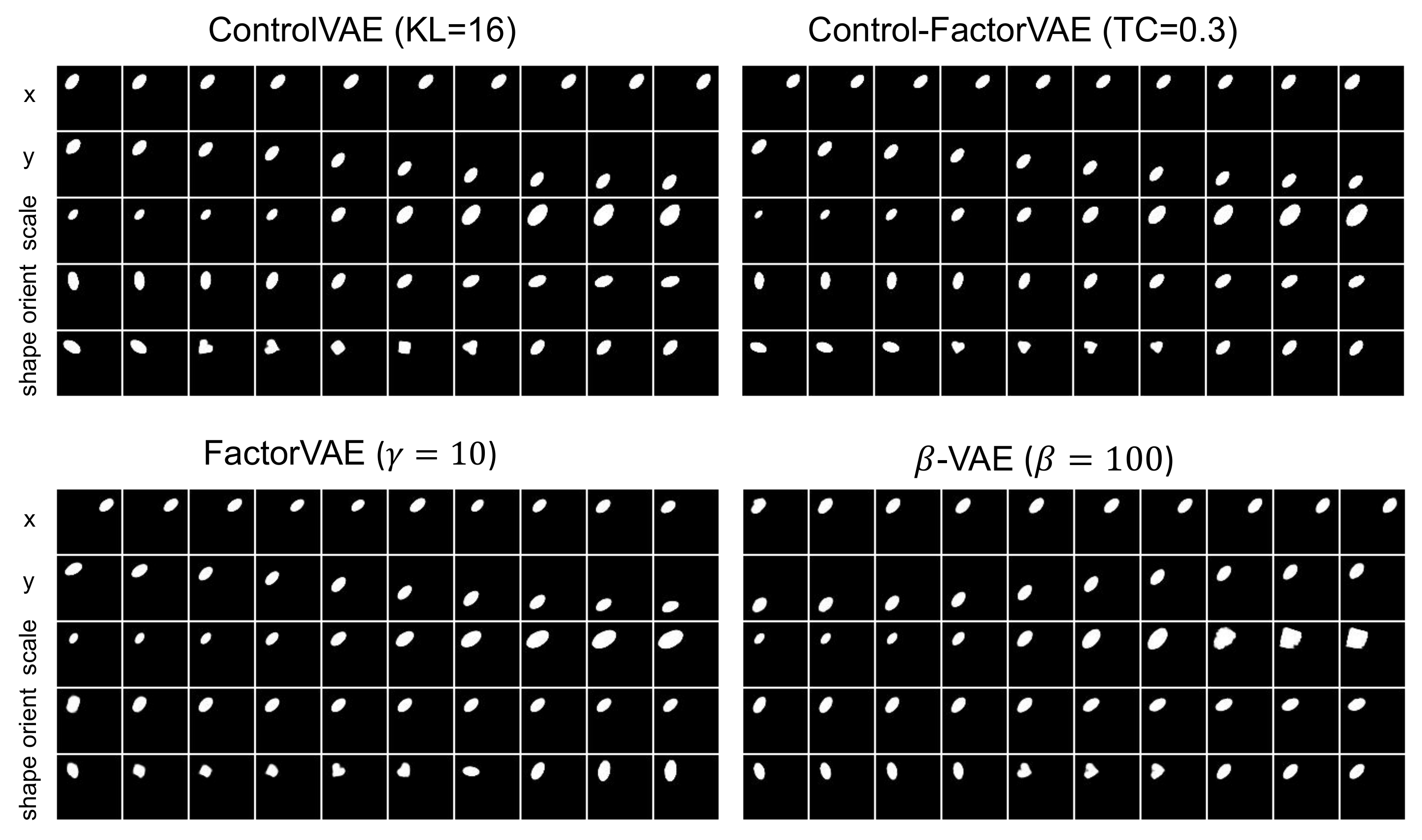}}
\caption{Rows: latent traversals ordered by the value of KL-divergence with the prior in a descending order. Following work~\cite{burgess2018understanding}, we initialize the latent representation from a seed image, and then traverse a single latent dimension in a range of $[-3,3]$, while keeping the remaining latent dimensions fixed. ControlVAE and Control-FactorVAE can disentangle all the five generative factors for 2D Shapes data, while $\beta$-VAE entangles the scale and shape (in 3rd row) and FactorVAE does not disentangle $y$ position very well.}
\label{fig:sprites_image}
\end{center}
\end{figure*}

Since there does not exist an exactly accurate metric to measure disentanglement, we also show the qualitative results of different models in Fig.~\ref{fig:sprites_image}. We can observe that ControlVAE and Control-FactorVAE can discover all the five generative factors: positional latent ($x$ and $y$), scale, orientation and shape. However, $\beta$-VAE ($\beta=100$) disentangles four generative factors except for entangling the scale and shape together (in the third row), while FactorVAE ($\gamma=10$) does not disentangle position $y$ very well in Fig.~\ref{fig:sprites_image}. Based on the above experimental results, we can conclude that ControlVAE and Control-FactorVAE achieve a better reconstruction quality than the baselines for the comparable disentanglement.

\subsection{Evaluation on ELBO and Reconstruction}
\begin{figure*}[!tb]
\centering     
\subfigure[ELBO]{\includegraphics[width=0.32\textwidth]{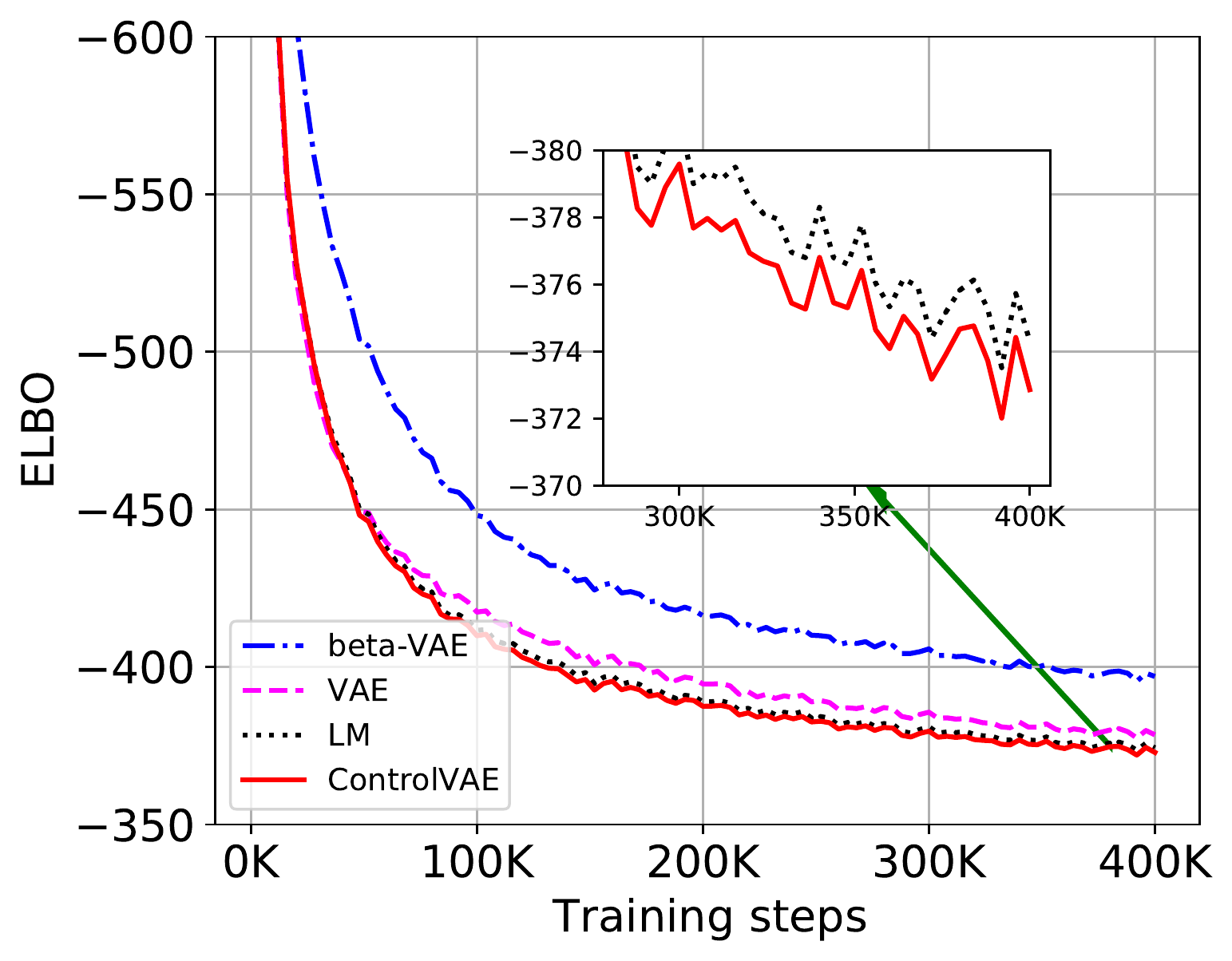}}
\subfigure[Reconstruction loss]{\includegraphics[width=0.32\textwidth]{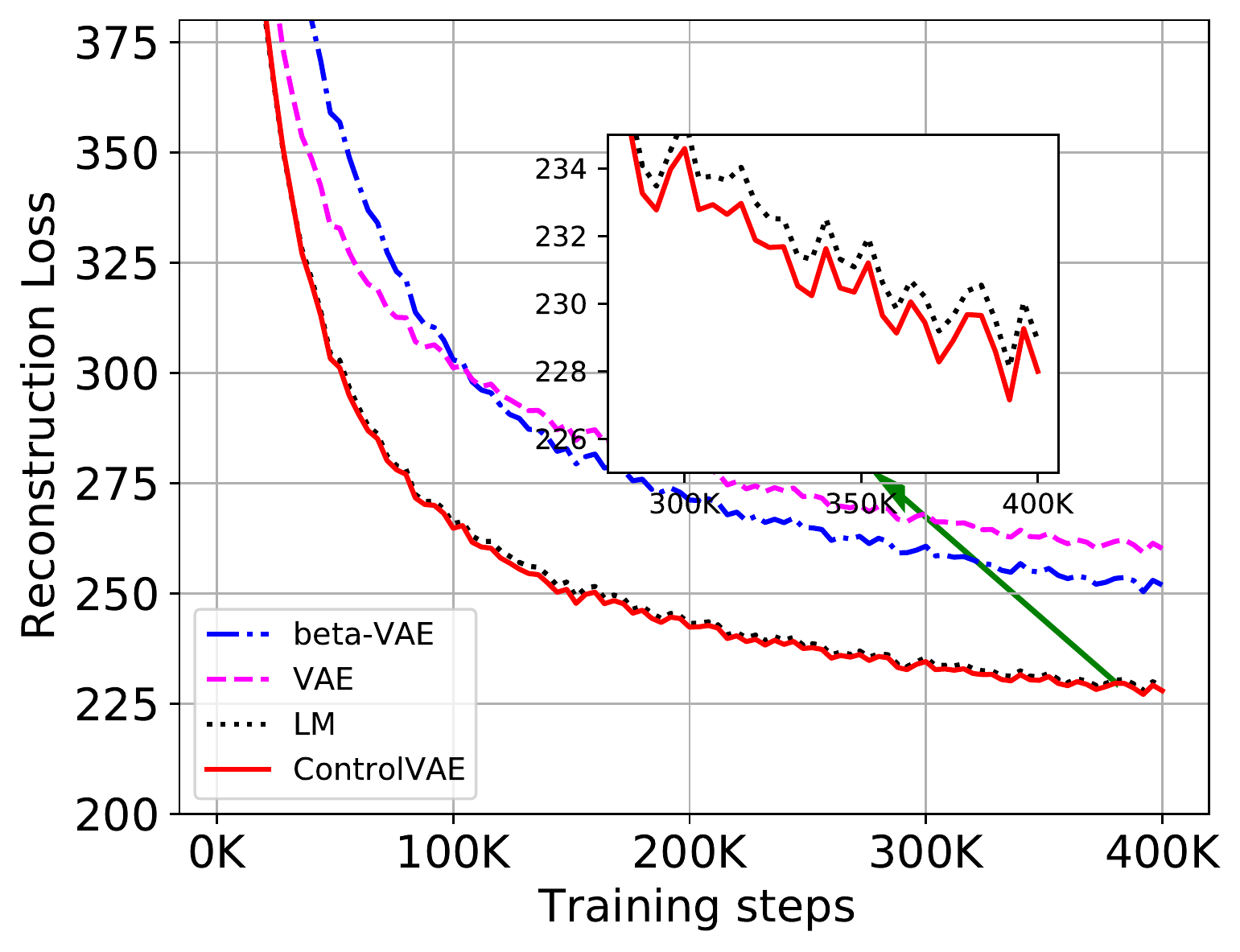}}
\subfigure[KL-divergence]{\includegraphics[width=0.32\textwidth]{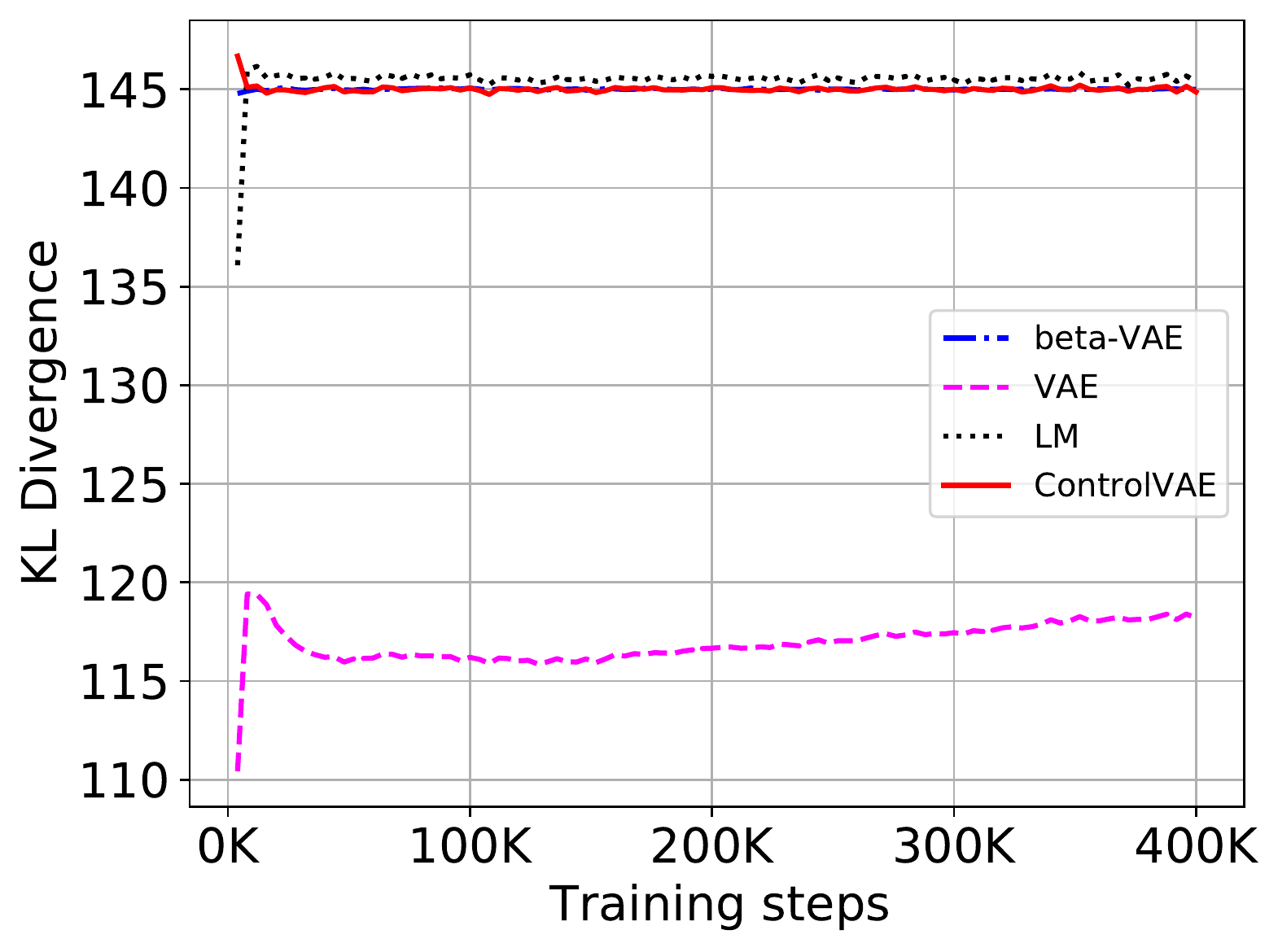}}
\caption{Performance comparison for different methods on the CIFAR-10 averaged over $5$ random seeds. Fig.(a)(b) shows that ControlVAE has a higher ELBO and lower reconstruction loss than the other methods given the desired KL-divergence $145$. Fig.(c) illustrates that ControlVAE is able to stabilize the KL-divergence to the target value, 145, while Lagrange multiplier (LM) method has a bias so that it cannot stabilize the KL-divergence.}\label{fig:CelebA-compare}
\end{figure*}

{\color{black}
We also demonstrate that ControlVAE can change the optimization trajectory to improve the ELBO and the reconstruction quality over the basic VAE for image generation task. Here we follow the set point guidelines in Section~\ref{sec:bound} to choose the set point of KL-divergence, KL=145, to conduct experiments. We compare our method with the following baselines.
\begin{itemize}
\item \textbf{Lagrange multiplier (LM)}: it formulates the KL-divergence in the VAE objective as a constrained optimization problem using Lagrangian multiplier as mentioned in Section~\ref{sec:connect}.
\item \textbf{$\beta$-VAE}~\cite{burgess2018understanding}: it assigns a large weight on the KL term to force the value of KL-divergence to be stabilized to a specified value.
\item \textbf{VAE}: It is a basic VAE model which consists of reconstruction term and KL-divergence term without any weight or constraint.
\end{itemize}

Fig.~\ref{fig:CelebA-compare}~(a) and (b) show the comparison of reconstruction error and ELBO under different set points of KL-divergence on CIFAR-10 dataset using $5$ different random seeds during model training. We can observe from it that ControlVAE-KL (KL=145) has the highest ELBO and lowest reconstruction error among them. This is because ControlVAE can achieve a good trade-off between reconstruction quality and KL-divergence to improve the optimization trajectory. The ELBO of Lagrange multiplier (LM) is slightly lower than the ControlVAE, because it suffers from local minima caused by the non-linear term. Moreover, ControlVAE outperforms $\beta$-VAE ($\beta$=30) in~\cite{burgess2018understanding}, because it uses dynamic learning to stabilize the KL-divergence while the latter assigns a large and fixed hyper-parameter to the KL term. We also compare the stability performance of different method as illustrated in Fig.~\ref{fig:CelebA-compare}~(c). It can be observed that ControlVAE and $\beta$-VAE can stabilize the KL-divergence to a target value, while LM has a bias to its target value. In other words, LM is unable to precisely control the KL divergence to a specified value.

We further use FID~\cite{lucic2018gans} and ELBO to evaluate the performance of ControlVAE using testing datasets, CIFAR-10 and CelebA, as illustrated in Table~\ref{tab:cifar10} and \ref{tab:celeba}. It can be observed from them that ControlVAE outperforms the other methods in terms of FID and ELBO in the testing. Therefore, ControlVAE can improve the optimization trajectory and the reconstruction quality for image generation via choosing the desired value of KL-divergence from our derived set points.
}

\begin{table}
\caption{Performance comparison for different methods using ELBO and FID on CIFAR-10 data set averaged over $5$ random seeds.}
\label{tab:cifar10}
\centering
\begin{tabular}{lllll}
	\toprule
	Methods/metric &  ELBO & FID\\
	\midrule
	VAE (KL=118)   & -372.10 $\pm$ 0.64 & 135.25 $\pm$ 0.31 \\
	ControlVAE (KL=145) & \textbf{-365.18 $\pm$ 0.37} & \textbf{122.36 $\pm$ 0.33}  \\
	LM (KL=145) & -366.87 $\pm$ 0.90 & 124.76 $\pm$ 0.58 \\ 
	$\beta$-VAE (KL=145)  & -388.82 $\pm$ 0.76 & 134.05 $\pm$ 0.54   \\
	\bottomrule
	\end{tabular}

\end{table}

\begin{table}[htb]
\caption{Performance comparison for different methods using ELBO and FID on CelebA data set using $5$ random seeds.}
\label{tab:celeba}
\centering
\begin{tabular}{lllll}
\toprule
Methods/metric &  ELBO & FID\\
\midrule
VAE (KL=127)   & -472.21 $\pm$  0.52 & 89.33 $\pm$ 0.51 \\
ControlVAE (KL=155) & \textbf{-468.93 $\pm$ 0.71} & \textbf{86.91 $\pm$ 0.49}  \\
LM (KL=155) & -469.09 $\pm$ 0.85 & 87.01 $\pm$ 0.37 \\
$\beta$-VAE (KL=155)  & -494.06 $\pm$ 0.80 & 90.27 $\pm$ 0.40    \\
\bottomrule
\end{tabular}
\end{table}

\subsection{Evaluation on Language Modeling}
Finally, we compare the performance of ControlVAE with the following baselines for mitigating KL vanishing in text generation~\cite{bowman2015generating}. 
\begin{itemize}
\item \textbf{Cost annealing}~\cite{bowman2015generating}: This method gradually increases the hyperparameter on KL-divergence from $0$ until to $1$ after $N$ training steps using sigmoid function.
\item  \textbf{Cyclical annealing}~\cite{liu2019cyclical}: This method splits the training process into $M$ cycles and each increases the hyperparameter from $0$ until to $1$ using a linear function.
\end{itemize}

\begin{figure*}[!thb]
\centering     
\subfigure[KL divergence]{\includegraphics[width=0.4\textwidth]{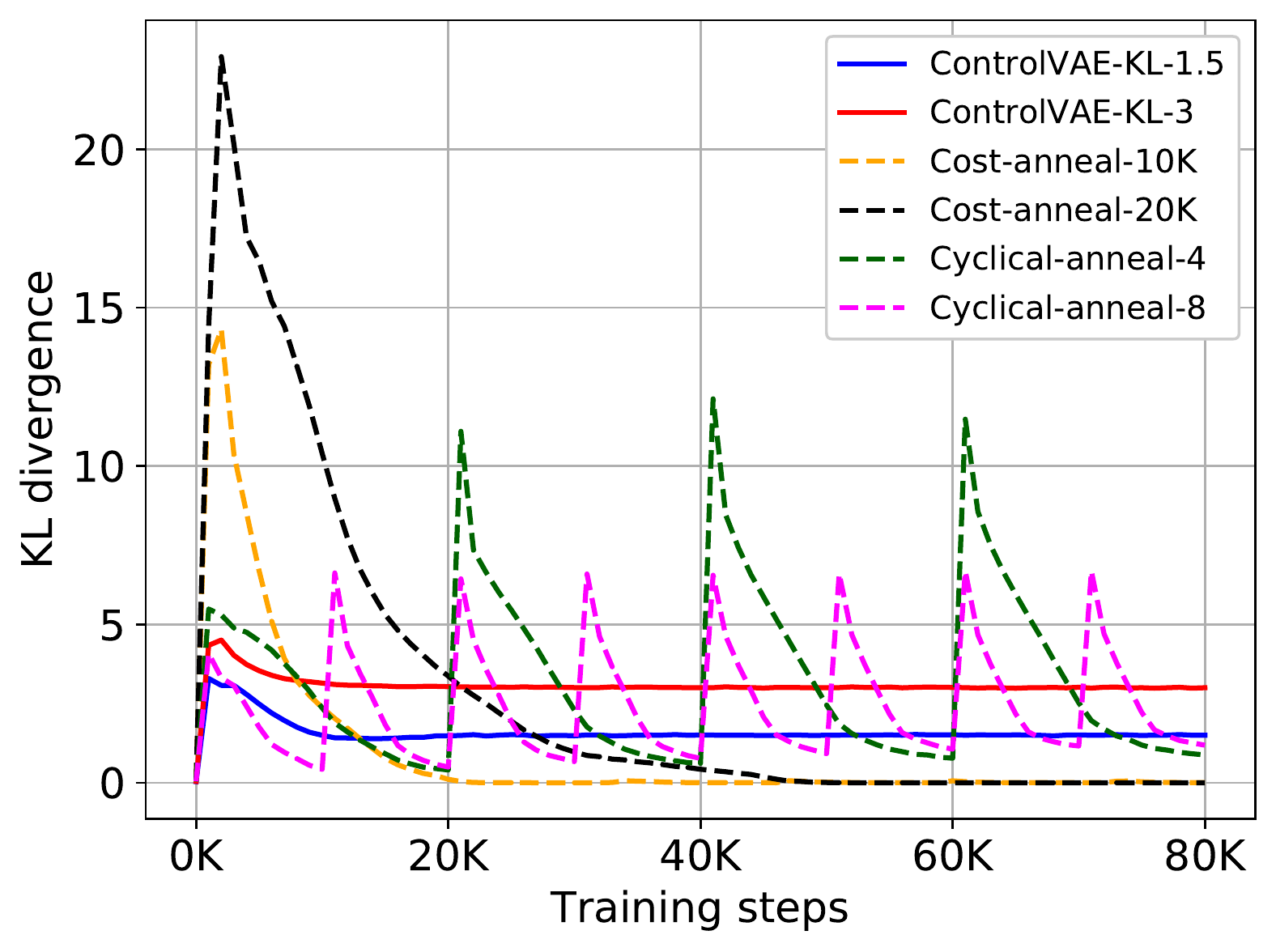}}
\subfigure[Reconstruction loss]{\includegraphics[width=0.4\textwidth]{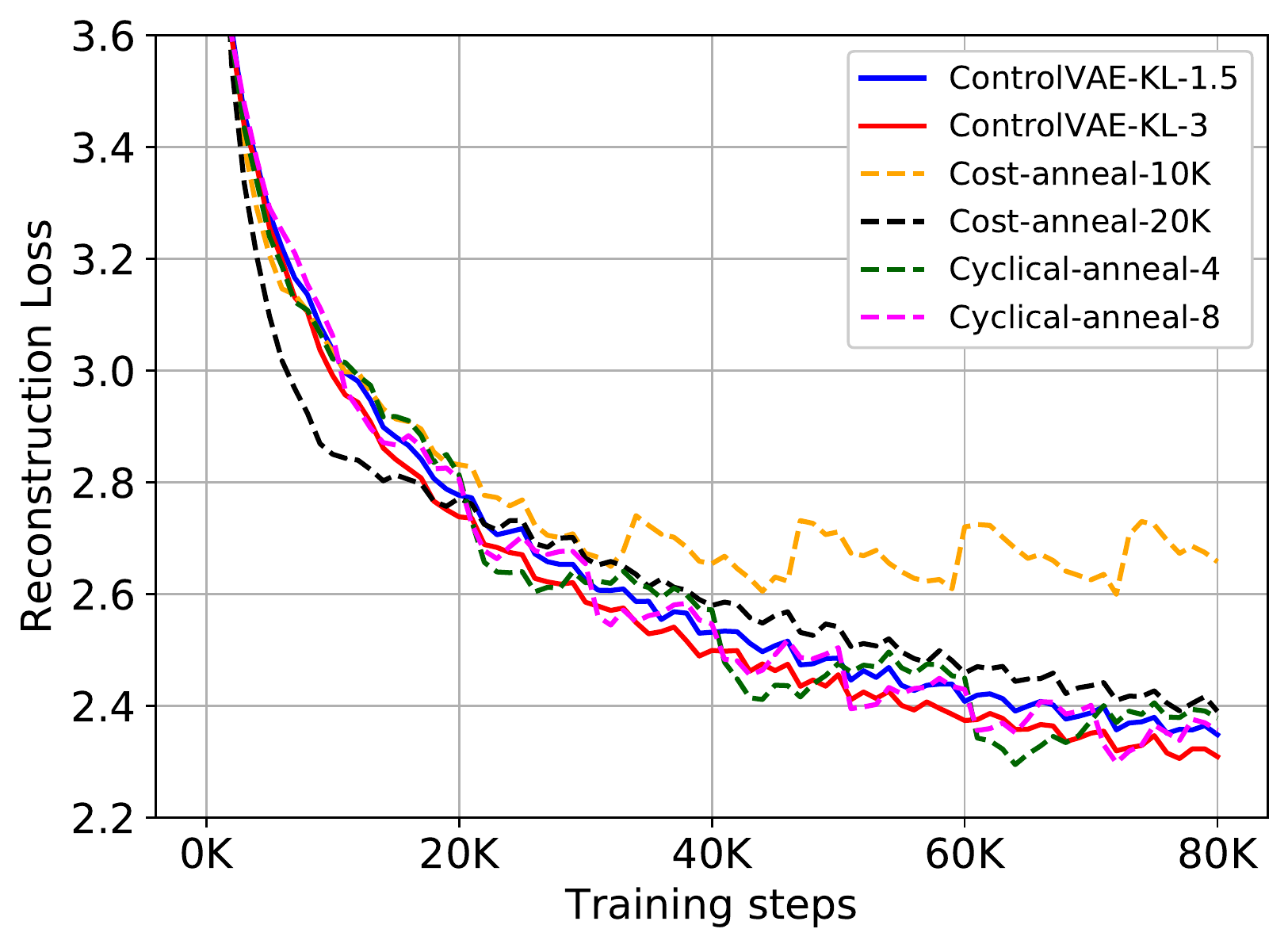}}
\subfigure[ELBO]{\includegraphics[width=0.4\textwidth]{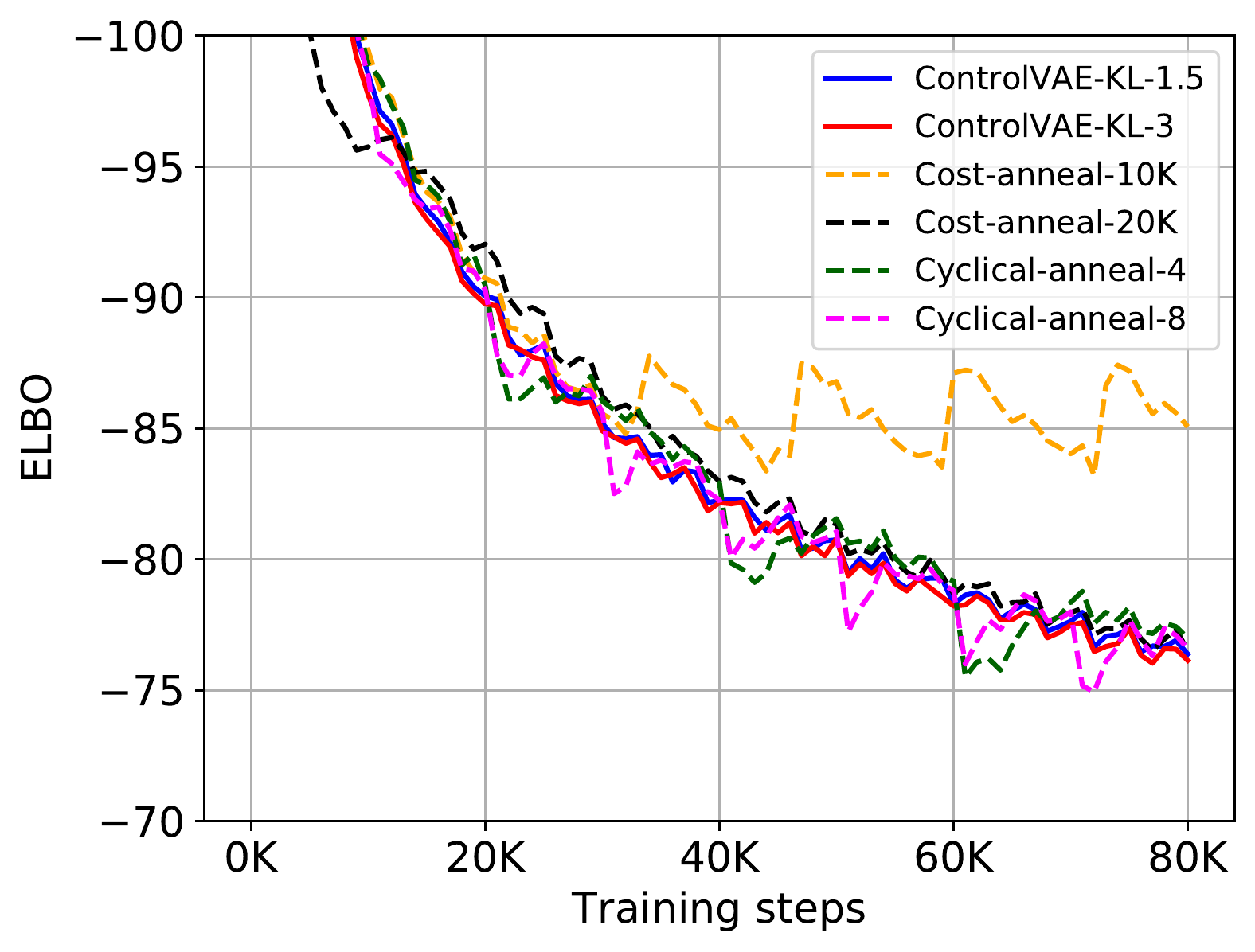}}
\subfigure[$\beta(t)$]{\label{fig:c}\includegraphics[width=0.4\textwidth]{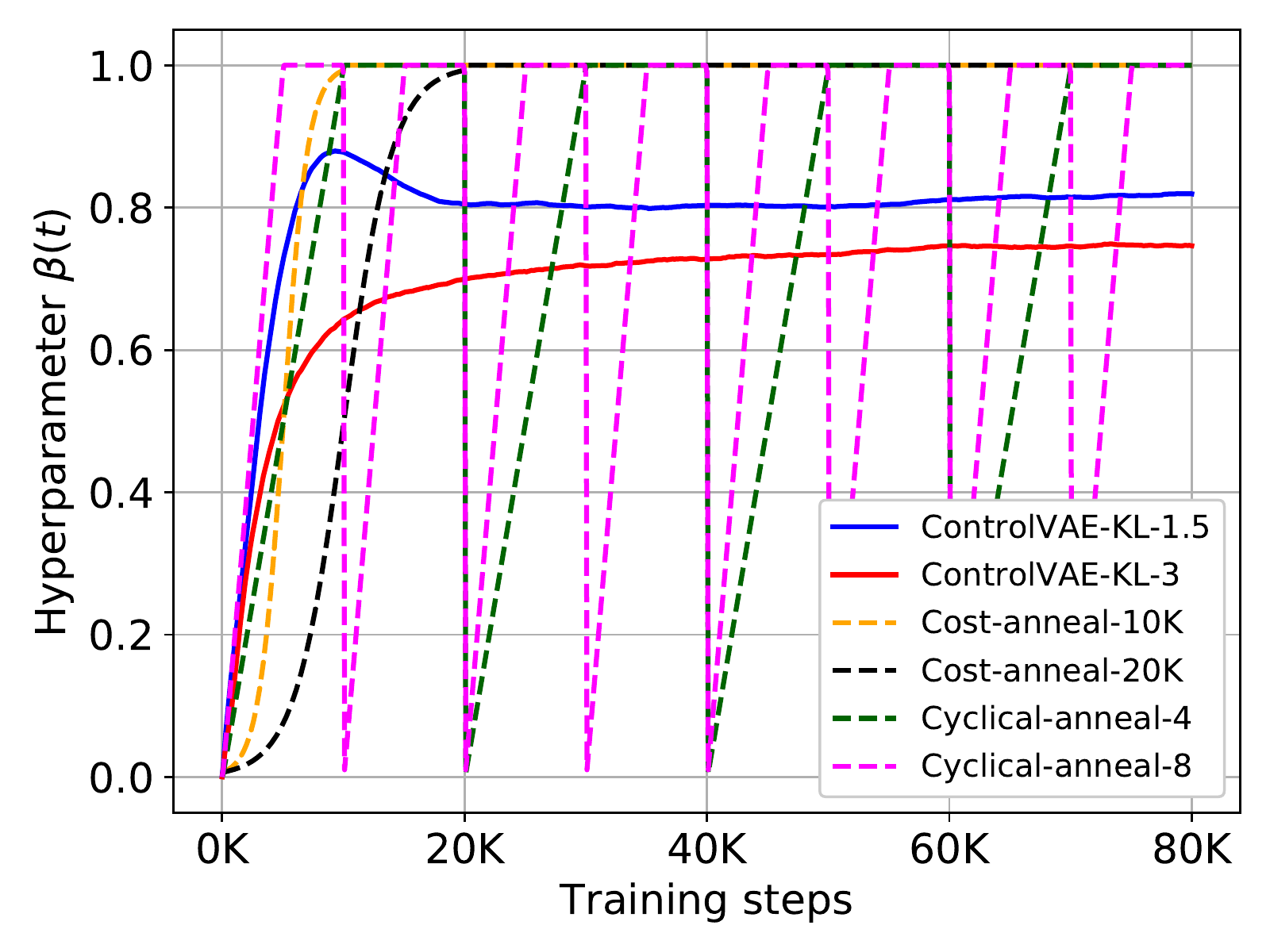}}
\vskip -0.02in
\caption{Performance comparison for different methods on the PTB data. (a) shows that ControlVAE and Cyclical annealing ($4, 8$ cycles) can avert KL vanishing, while Cost annealing still suffers from KL vanishing after $20K$ and $50K$ training steps. Moreover, ControlVAE can control the KL-divergence and also has lower reconstruction errors than the other methods in (b). Fig. (c) illustrates that ControlVAE has higher ELBO than the baselines.}\label{fig:KL-vanish}
\end{figure*}

\begin{table*}[!tb]
\caption{Performance comparison for different methods on dialog-generation using SW data over $5$ random seeds. Dis-$n$: higher is better. PPL: lower is better, and self-BLEU lower is better.}
\label{tab:dist}
\begin{center}
\begin{small}
\begin{tabular}{llllll}
\toprule
Methods/metric & Dis-1 & Dis-2 & self-BLEU-2 & self-BLEU-3 & PPL \\
\midrule
ControlVAE-KL-35  & \textbf{6.27K} $\pm$ 41 &  \textbf{95.86K} $\pm$ 1.02K & \textbf{0.663} $\pm$  0.012 & \textbf{0.447} $\pm$ 0.013 & \textbf{8.81} $\pm$ 0.05  \\
ControlVAE-KL-25 & 6.10K $\pm$ 60 & 83.15K $\pm$ 4.00K & 0.698 $\pm$ 0.006 & 0.495 $\pm$  0.014  &  12.47 $\pm$ 0.07 \\
Cost anneal-KL-17  & 5.71K $\pm$ 87 & 69.60K $\pm$ 1.53K & 0.721 $\pm$ 0.010 & 0.536 $\pm$ 0.008 & 16.82 $\pm$ 0.11 \\
Cyclical (KL = 21.5)  & 5.79K $\pm$ 81 & 71.63K $\pm$ 2.04K &  0.710 $\pm$ 0.007 & 0.524 $\pm$  0.008 & 17.81 $\pm$ 0.33 \\
\bottomrule
\end{tabular}
\end{small}
\end{center}
\end{table*}

Fig.~\ref{fig:KL-vanish} illustrates the comparison results of KL divergence, reconstruction loss and hyperparamter, $\beta(t)$, for different methods on the PTB dataset. Note that, here ControlVAE-KL-$v$ means we set the KL-divergence to a desired value $v$ (e.g., 3) for our PI controller following the set point guidelines in Section~\ref{sec:bound}. Cost-annealing-$v$ means we gradually increase the hyperparameter, $\beta(t)$, from $0$ until to $1$ after $v$ steps using sigmoid function. We observe from Fig.~\ref{fig:KL-vanish}(a) that ControlVAE (KL=1.5, 3) and Cyclical annealing ($4, 8$ cycles) can avert the KL vanishing. However, our ControlVAE is able to stabilize the KL-divergence while cyclical annealing could not. Moreover, our method has a lower reconstruction loss than the cyclical annealing in Fig.~\ref{fig:KL-vanish} (b). Cost annealing method still suffers from KL vanishing, because we use the Transformer~\cite{vaswani2017attention} as the decoder, which can predict the current data based on previous ground-truth data. In addition, we can observe from Fig.~\ref{fig:KL-vanish} (c) that ControlVAE improves the ELBO over the baselines, which means it can change the optimization trajectory.  Fig.~\ref{fig:KL-vanish} (d) illustrates the tuning result of $\beta(t)$ by ControlVAE compared with other methods. We can discover that our $\beta(t)$ gradually converges to around a certain value. Note that, here $\beta(t)$ of ControlVAE does not converge to $1$ because we slightly increase the value of KL-divergence (produced by the original VAE) in order to improve the diversity of generated data.

In order to further demonstrate ControlVAE can improve the diversity of generated text, we apply it to dialog-response generation using the Switchboard(SW) dataset. Following~\cite{zhao2017learning}, we adopt a conditional VAE~\cite{zhao2017learning} that generates dialog conditioned on the previous response. We use metric $Dis$-$n$~\cite{xu2018dp} and self-BLEU~\cite{zhu2018texygen} (with 1000 sampled results) to measure the diversity of generated data, and perplexity (PPL)~\cite{jelinek1977perplexity} to measure how well the probability distribution predicts a sample. Table~\ref{tab:dist} illustrates the comparison results for different approaches. We can observe that ControlVAE has more distinct grams and lower self-BLEU than the baselines when the desired KL-divergence is set to $35$ and $25$. In addition, it has lower PPL than the other methods. Thus, we can conclude that ControlVAE can improve the diversity of generated data and generation performance. We also illustrate some examples of generated dialog by ControlVAE in Appendix~\ref{app:exp-dialog}.

%% file: ablation.tex
\section{Ablation Studies}
In this section, we conduct ablation studies to study the impact of hyper-parameters on the performance of the proposed ControlVAE.

\subsection{Effect of Set Points of KL-divergence on Disentanglement}
We first study the influence of different target values, $C$, of KL-divergence on the disentanglement representation learning. We change the target KL-divergence $C$ from 16 to 19 with step 1 while keeping the other parameters unchanged. Table~\ref{tab:abl-KL} illustrates the RMIG score of five disentangled factors and the overall score. We can observe from it that ControlVAE has the highest RMIG score on average when $KL=16$, since it disentangles three factors: position $x$, $y$ and shape, better than the other KL values. In addition, when the target KL-divergence is too large (e.g., $KL=19$), it may hurt the performance of disentanglement. This is because when the KL-divergence is increased, multiple latent factors may transmit through the information channels together.

\begin{table}[!htb]
\caption {RMIG score for different target KL-divergence averaged over 5 random seeds. The higher is better.}\label{tab:abl-KL}
\vskip -0.1in
\begin{center}
\begin{small}
\begin{tabular}{lcccc}
\toprule
Metric & KL=16 & KL=17 & KL=18 & KL=19  \\
\midrule
pos. $x$ &\textbf{ 0.7490} & 0.7477 & 0.7229 & 0.7136  \\
pos. $y$ & \textbf{0.7413} & 0.7324 & 0.7221 & 0.7074 \\
Shape & \textbf{0.0958} & 0.0795 & 0.0701 & 0.0721 \\
Scale &  0.6288 & 0.6206 & \textbf{0.6404} &  0.6035 \\
Orientation & 0.0178 & 0.0177 & \textbf{0.0582} &  0.0126 \\
RMIG & \textbf{0.4637} & 0.4396 & 0.4427 & 0.4218 \\
\bottomrule
\end{tabular}
\end{small}
\end{center}
\end{table}

\subsection{Effect of Step Value on Disentanglement}
For disentanglement representation learning, ControlVAE uses an annealing method to increase the target KL-divergence $C$ with step $s$ every $K = 5,000$ iterations. Hence, we try to learn how the step value, $s$, impacts the disentanglement. Table~\ref{tab:abl-step} shows the comparison results of RMIG under different step values. It can be seen that when the step value $s=0.15$, ControlVAE has the best disentanglement performance. Besides, it can better disentangle the position $x$ factor and orientation when $s=0.05$ and $s=0.2$, respectively. We also find that the RMIG score of ControlVAE may decrease if the step value is too large, due to the fact that multiple factors would be entangled together with a large KL-divergence (information bottleneck).

\begin{table}[!ht]
\caption {RMIG score for different step values averaged over 5 random seeds. The higher is better.}
\label{tab:abl-step}
\vspace{-0.15in}
\begin{center}
\begin{small}
\begin{tabular}{lcccc}
\toprule
Metric & $s=0.05$ & $s=0.1$ & $s=0.15$ & $s=0.2$  \\
\midrule
pos. $x$ & \textbf{0.7520} & 0.7391 & 0.7490 & 0.7210 \\
pos. $y$ & 0.7317 & 0.7405 & \textbf{0.7413}  & 0.7205 \\
Shape & 0.0749 & 0.0488 & \textbf{0.0958} & 0.0340 \\
Scale &  0.5695 &  0.5916 & \textbf{0.6288} &  0.6159\\
Orientation & 0.0551 & 0.0467 & 0.0178 & \textbf{0.0608} \\
RMIG & 0.4366 & 0.433 & \textbf{0.4637} & 0.4304 \\
\bottomrule
\end{tabular}
\end{small}
\end{center}
\end{table}

\subsection{Effect of Batch Size on ELBO}
Next, we study how the batch size of model training influences the ELBO on the image generation task. In our experiment, we change batch size from $50$ to $150$ to evaluate the performance of ControlVAE on the CIFAR10 data set. Table~\ref{tab:abl-batchsize} illustrates the ELBO of ControlVAE under different batch sizes. It can be observed from this table that the ELBO of ControlVAE is higher when trained with a large bath size than that with small batch size, $50$. The main reason is that the output KL divergence is not very stable when the bath size is small during model training. Hence, we need to use a large batch size, such as 100, to train our model.

\begin{table}[!htb]
\caption {ELBO of ControlVAE under different batch sizes averaged over 5 random seeds. The higher is better.}
\label{tab:abl-batchsize}
\vspace{-0.15in} 
\begin{center}
\begin{small}
\begin{tabular}{lcccc}
\toprule
Metric & batch =50 & batch =100 & batch =150  \\
\midrule
ELBO & -375.99 $\pm$ 1.82 & \textbf{-368.18 $\pm$ 1.00} & -370.23 $\pm$ 0.82 \\
\bottomrule
\end{tabular}
\end{small}
\end{center}
\end{table}

\subsection{Effect of Embedding Size on ELBO}
We also study the influence of embedding size of latent space, $z$, on the ELBO for image generation task. In this paper, the embedding size, 100, 200, and 500 are used to conduct experiments on the CIFAR10 data set. In Table~\ref{tab:abl-embed}, we can see that when the embedding size of latent variable $z$ is set to $200$, ControlVAE has the highest ELBO.

\begin{table}[!htb]
\caption {ELBO comparison under different embedding sizes of latent variable over 5 random seeds. The higher is better.}
\label{tab:abl-embed}
\vspace{-0.15in}
\begin{center}
\begin{small}
\begin{tabular}{lcccc}
\toprule
Metric & $z$ = 100 & $z$ = 200 & $z$ = 500  \\
\midrule
ELBO & -365.18 $\pm$ 0.37 & \textbf{-361.80 $\pm$ 0.86} & -364.17 $\pm$ 1.25 \\
\bottomrule
\end{tabular}
\end{small}
\end{center}
\end{table}

%% file: relatedwork.tex
\section{Related Work}
\label{sec:relatedwork}

There are many work involving a trade-off between reconstruction and KL-divergence for VAEs applications. For disentangled representation learning, researchers proposed $\beta$-VAE ($\beta > 1$)~\cite{higgins2017beta,burgess2018understanding} that assigns a large and fixed hyperparameter, $\beta$, to put more emphasis on the KL divergence to encourage disentangled latent representations. It, however, sacrifice the reconstruction quality in order to obtain better disentangling. Then some follow-up work~\cite{chen2018isolating,kim2018disentangling} further factorize the KL-divergence term to improve the reconstruction quality. However, these methods still assign a fixed and large hyperparameter to the decomposed terms in the objective, resulting in high reconstruction error. In contrast, ControlVAE dynamically tunes $\beta$ during optimization to achieve better disentangling \textit{and} reconstruction quality.

In order to improve the sample generation quality of VAEs~\cite{dai2019diagnosing,xiao2019generative,ghosh2019variational,alemi2017fixing,zhao2019infovae}, some researchers tried to reduce the weight of KL-divergence to make the decoder produce sharper outputs. Though they can obtain impressive sample quality, they suffer severely from the trade-off in the way that the latent distribution is far away from the prior. Recent studies adopted a constrained optimization for reconstruction error~\cite{rezende2018taming,klushyn2019learning} to achieve the trade-off between reconstruction error and KL-divergence. They may suffer from posterior collapse if the inference network fails to cover the latent space while our can totally avert posterior collapse. Moreover, different from their work, we try to optimize KL-divergence (information bottleneck) as a constraint. Our method and theirs complement each other for different applications.

In language modeling, VAE often suffers from KL vanishing, due to a powerful decoder, such as Transformer~\cite{vaswani2017attention} and LSTM. To remedy this issue, one popular way is to add a hyperparameter $\beta$ on the KL term~\cite{bowman2015generating,liu2019cyclical}, and then gradually increases it from $0$ until $1$. However, the existing methods~\cite{yang2017improved,bowman2015generating,liu2019cyclical}, such as KL cost annealing and cyclical annealing, cannot totally solve KL vanishing or explicitly control the value of KL-divergence since they blindly change $\beta$ without observing the actual KL-divergence during model training. Conversely, our approach can avert KL vanishing and stabilize the KL-divergence to a desired value.

%% file: conclusion.tex
\section{Conclusion}
\label{sec:conclusion}
In this paper, we proposed a general controllable VAE framework, ControlVAE, that combines automatic control with the basic VAE framework to improve the performance of the VAE models. A novel non-linear PI controller was designed to control the value of KL divergence during model training. We also developed a new variant of ControlVAE, Control-FactorVAE, to improve the disentanglement. In addition, simplified theoretical analysis was provided to help choose the set points of KL-divergence for our method. Then we presented the connections between ControlVAE and other models. The evaluation results showed that ControlVAE is able to improve the ELBO over the basic VAE by changing its optimization trajectory on the task of image generation. For disentangled representation learning, it significantly improves the reconstruction quality while achieving a comparable disentanglement with the best baselines. We also demonstrated that ControlVAE can totally avert the KL-vanishing (posterior collapse) problem and control the diversity of generated data for language modeling.

%% file: appendix.tex
\appendices

%
%
%

\section{Proof of Theorem 1}
\label{app:theorem1}
We are going to prove the Theorem~\ref{the:z_eq} below.
\begin{proof}
Assuming the decoder to be $k$-Lipschitz continuous, we have
\begin{equation} \label{eq:lipt}\small
\mathbb{E}_{q_\phi(\mathbf{z}|\mathbf{x})} [\log p_\theta(\mathbf{x|z})] - \mathbb{E}_{q_{\phi'}(\mathbf{z_0|x)}} [\log p_\theta(\mathbf{x|z}_0)] \leq k \|E[\mathbf{z}] - E[\mathbf{z}_0] \|,
\end{equation}
where $k$ is the Lipschitz constant, and we can assume $k=1$ for simplicity, since this is achievable if spectral normalization ~\cite{miyato2018spectral} is used. 

Our next step is to compute the bound of $\|E[\mathbf{z}] - E[\mathbf{z}_0] \|$. Both $\mathbf{z}$ and $\mathbf{z_0}$ follow Gaussian distributions with diagonal covariance, $q(\mathbf{z_0}|\mathbf{x}) \equiv \mathcal{N}(\boldsymbol{\mu}_1,\boldsymbol{\sigma}_1)$ and $q(\mathbf{z}|\mathbf{x}) \equiv \mathcal{N}(\boldsymbol{\mu}_2,\boldsymbol{\sigma}_2)$, respectively. Namely, they can be denoted by $\mathbf{z_0}= \boldsymbol{\mu}_1 + \epsilon \boldsymbol{\sigma}_1$ and $\mathbf{z}= \boldsymbol{\mu}_2 + \epsilon \boldsymbol{\sigma}_2$, where $\epsilon \in \mathcal{N}(0,I)$. The prior distribution is always unit Gaussian: $p(\mathbf{z}) \sim \mathcal{N}(0,I)$. According to ~\cite{kingma2013auto}, the KL-divergence term of the basic VAE can be expressed as
\begin{equation}\label{eq:vae-KL}
KL_{vae} = \frac{1}{2} \sum_{n=1}^{N}\big( \mu_{1,n}^2 + \sigma_{1,n}^2 -1 - \log \sigma_{1,n}^2 \big)
\end{equation}
It is easy to prove that $\sigma_{1,n}^2 -1 - \log \sigma_{1,n}^2 \geq 0$:
\begin{proof}
Let $\sigma_{1,n}^2$ be $x$, then we need to prove $x-1-\log x \geq 0$. Let $g(x) = x-1-\log x$. When $x=1$, $g(x)$ can get the minimal value, $g(1)=0$. Thus, we can obtain $\sigma_{1,n}^2 -1 - \log \sigma_{1,n}^2 \geq 0$.
\end{proof}
Therefore, Eq.~\eqref{eq:vae-KL} can be expressed as
\begin{equation}\label{eq:vae-KL-simple}
KL_{vae} \geq \frac{1}{2} \sum_{n=1}^{N} \mu_{1,n}^2 = \frac{1}{2} ||\boldsymbol{\mu}_1||^2
\end{equation}

Similarly, for the KL-divergence of ControlVAE, we have
\begin{equation}\label{eq:cvae-KL}
KL_{cvae} = KL_{vae}+d \geq  \frac{1}{2} \sum_{n=1}^{N} \mu_{2,n}^2 =  \frac{1}{2} ||\boldsymbol{\mu}_2||^2.
\end{equation}
Adding Eq.~\eqref{eq:vae-KL-simple} to \eqref{eq:cvae-KL}, we can have
\begin{equation}\label{eq:cvae-KL2}
||\boldsymbol{\mu}_1||^2 + ||\boldsymbol{\mu}_2||^2 \leq 4KL_{vae} + 2d
\end{equation}

According to norm squared inequality, we can have
\begin{equation}\label{eq:cvae-KL-inequal}
||\boldsymbol{\mu}_1 - \boldsymbol{\mu}_2||^2 \leq
2||\boldsymbol{\mu}_1||^2 + 2||\boldsymbol{\mu}_2||^2 \leq 8KL_{vae} + 4d
\end{equation}
Since $\boldsymbol{\mu}_1$ and $\boldsymbol{\mu}_2$ denote the mean of latent variables $\boldsymbol{z_0}$ and $\boldsymbol{z}$, we have
\begin{equation}\label{eq:exp_z_z0}
|| E[\boldsymbol{z_0}] - E[\boldsymbol{z}] || = ||\boldsymbol{\mu}_1 - \boldsymbol{\mu}_2|| \leq \sqrt{4 (2KL_{vae} + d)}.
\end{equation}
Hence, based on~\eqref{eq:lipt} and \eqref{eq:exp_z_z0}, we can get the following bound
\begin{equation}
\mathbb{E}_{q_\phi(\mathbf{z}|\mathbf{x})} [\log p_\theta(\mathbf{x|z})] - \mathbb{E}_{q_\phi(\mathbf{z_0|x)}} [\log p_\theta(\mathbf{x|z}_0)] \leq \sqrt{4 (2KL_{vae} + d)}.
\end{equation}
\end{proof}


\section{Model Configurations and hyperparameter settings}
\label{sec:configure}
We summarize the detailed model configurations and hyperparameter settings for ControlVAE in the following three applications: language modeling, disentanglement representation learning and image generation.

\subsection{Experimental Details for Disentangling}
Following the same model architecture of $\beta$-VAE~\cite{higgins2017beta}, we adopt a convolutional layer and deconvolutional layer for our experiments. We use Adam optimizer with $\beta_1=0.90$, $\beta_2=0.99$ and a learning rate tuned from $10^{-4}$. We set $K_p$ and $K_i$ for PI algorithm to $0.01$ and $0.001$, respectively. For the step function, we set the step, $s$, to $0.15$ per $K=5,000$ training steps as the information capacity (desired KL- divergence) increases from $0.5$ until $18$ for 2D Shape data. ControlVAE uses the same encoder and decoder architecture as $\beta$-VAE except for plugging in PI control algorithm, illustrated in Table~\ref{tab:2Dshape_model}.

\begin{table}[htb]
\caption{Encoder and decoder architecture for disentangled representation learning on 2D Shapes data.}
\label{tab:2Dshape_model}
\begin{center}
\begin{scriptsize}
\begin{tabular}{|l|l|}
\toprule
Encoder & Decoder \\
\midrule
Input $64\times64$ binary image  & Input $\in \mathbb{R}^{10}$ \\
\midrule
$4\times4$ conv. $32$ ReLU. stride 2 &  FC. 256 ReLU.\\
\midrule
$4\times4$ conv. $32$ ReLU. stride 2 & $4\times4$ upconv. $256$ ReLU. \\
\midrule
$4\times4$ conv. $64$ ReLU. stride 2 &  $4\times4$ upconv. $64$ ReLU. stride 2.\\
\midrule
$4\times4$ conv. $64$ ReLU. stride 2 & $4\times4$ upconv. $64$ ReLU. stride 2 \\
\midrule
$4\times4$ conv. $256$ ReLU. &  $4\times4$ upconv. $32$ ReLU. stride 2 \\
\midrule
FC $256$. FC. $2 \times 10$ &  $4\times4$ upconv. $32$ ReLU. stride 2 \\
\bottomrule
\end{tabular}
\end{scriptsize}
\end{center}
\end{table}

\subsection{Experimental Details for Image Generation}
Similar to the architecture of $\beta$-VAE, we use a convolutional layer with batch normalization as the encoder and a deconvolutional layer with batch normalization for our experiments. We use Adam optimizer with $\beta_1=0.90$, $\beta_2=0.99$ and a learning rate $10^{-4}$ for CelebA data. The size of latent variable is set to $500$, because we find it has a better reconstruction quality than $200$ and $400$. In addition, we set the desired value of KL-divergence to $170$ (same as the original VAE), $180$, and $200$. For PI control algorithm, we set $K_p$ and $K_i$ to $0.01$ and $0.0001$, respectively. We also use the same encoder and decoder architecture as $\beta$-VAE above except that we add the batch normalization to improve the stability of model training, as shown in Table~\ref{tab:celeba_model}. 

\begin{table}[htb]
\caption{Encoder and decoder architecture for image generation on CelebA data.}
\label{tab:celeba_model}
\begin{center}
\begin{scriptsize}
\begin{tabular}{|l|l|}
\toprule
Encoder & Decoder \\
\midrule
Input $128\times 128 \times 3$ RGB image  & Input $\in \mathbb{R}^{500}$ \\
\midrule
$4\times4$ conv. $32$ ReLU. stride 2 &  FC. 256 ReLU.\\
\midrule
$4\times4$ conv. $32$ ReLU. stride 2 & $4\times4$ upconv. $256$ ReLU. stride 2 \\
\midrule
$4\times4$ conv. $64$ ReLU. stride 2 &  $4\times4$ upconv. $64$ ReLU. stride 2.\\
\midrule
$4\times4$ conv. $64$ ReLU. stride 2 & $4\times4$ upconv. $64$ ReLU. stride 2 \\
\midrule
$4\times4$ conv. $256$ ReLU. stride 2 &  $4\times4$ upconv. $32$ ReLU. stride 2 \\
\midrule
FC $4096$. FC.$2 \times 500$ &  $4\times4$ upconv. $32$ ReLU. stride 2 \\
\bottomrule
\end{tabular}
\end{scriptsize}
\end{center}
\vskip -0.1in
\end{table}

\subsection{Experimental Details for Language Modeling}
For text generation on PTB data, we build the ControlVAE model on the basic VAE model, as in~\cite{bowman2015generating}. We use one-layer LSTM as the encoder and a three-layer Transformer with eight heads as the decoder and a Multi-Layer Perceptron (MLP) to learn the latent variable~$\mathbf{z}$. The maximum sequence length for LSTM and Transformer is set to $100$, respectively. And the size of latent variable is set to $64$. Then we set the dimension of word embedding to $256$ and the batch size to $32$. In addition, the dropout is $0.2$ for LSTM and Transformer. Adam optimization with the learning rate $0.001$ is used during training. Following the tuning guidelines above, we set the coefficients $K_p$ and $K_i$ of P term and I term to $0.01$ and $0.0001$, respectively. Finally, We adopt the source code on Texar platform to implement experiments~\cite{hu2019texar}. 

For dialog-response generation, we follow the model architecture and hyperparameters of the basic conditional VAE in~\cite{zhao2017learning}. We use one-layer Bi-directional GRU as the encoder and one-layer GRU as the decoder and two fully-connected layers to learn the latent variable. In the experiment, the size of both latent variable and word embeddings is set to $200$. The maximum length of input/output sequence for GRU is set to $40$ with batch size $30$. In addition, Adam with initial learning rate $0.001$ is used. In addition, we set the same $K_p$ and $K_i$ of PI algorithm as text generation above.
The model architectures of ControlVAE for these two NLP tasks are illustrated in Table~\ref{tab:ptb_model},~\ref{tab:sw_model}.

\begin{table}[htb]
\caption{Encoder and decoder architecture for text generation on PTB data.}
\label{tab:ptb_model}
\begin{center}
\begin{small}
\begin{tabular}{|l|l|}
\toprule
Encoder & Decoder \\
\midrule
Input $n$ words $\times 256$ & Input $\in \mathbb{R}^{64}$, $n \times 256$ \\
\midrule
1-layer LSTM & FC $64 \times 256$ \\
\midrule
FC $64 \times 2$ & 3-layer Transformer 8 heads \\
\bottomrule
\end{tabular}
\end{small}
\end{center}
\end{table}

\begin{table}[htb]
\caption{Encoder and decoder architecture for dialog generation on Switchboard (SW) data.}
\label{tab:sw_model}
\begin{center}
\begin{small}
\begin{tabular}{|l|l|}
\toprule
Encoder & Decoder \\
\midrule
\textnormal{Input $n$ words $\times 200$} & \textnormal{Input $\in \mathbb{R}^{200}$}  \\
\midrule
\textnormal{1-layer bi-GRU} & \textnormal{FC $200 \times 400$} \\
\midrule
\textnormal{FC $200 \times 2$} & \textnormal{1-layer GRU} \\
\midrule
\textnormal{FC $200 \times 2$} & \\
\bottomrule
\end{tabular}
\end{small}
\end{center}
\vskip -0.1in
\end{table}

\section{Examples of Generated Dialog by ControlVAE}
\label{app:exp-dialog}
In this section, we show an example to compare the diversity and relevance of generated dialog by different methods, as illustrated in Table~\ref{tab:dialog_exp}. Alice begins with the open-ended conversation on choosing a college. Our model tries to predict the response from Bob. The ground truth response is ``um - hum''. We can observe from Table~\ref{tab:dialog_exp} that ControlVAE (KL=25, 35) can generate diverse and relevant response compared with the ground truth. In addition, while cyclical annealing can generate diverse text, some of them are not very relevant to the ground-truth response.

\begin{table}[!thb]
\caption{Examples of generated dialog for different methods. Our model tries to predict the response from Bob. The response generated by ControlVAE (KL=25,35) are relevant and diverse compared with the ground truth. However, some of reponse generated by cost annealing and cyclical annealing are not very relevant to the ground-truth data}
\vskip -0.1in
\label{tab:dialog_exp}
\begin{center}
\begin{small}
\begin{tabular}{p{0.225\textwidth}|p{0.225\textwidth}}
\toprule
\multicolumn{2}{p{0.44\textwidth}}{\textbf{Context}: (Alice) and a lot of the students in that home town sometimes $\langle$ unk $\rangle$ the idea of staying and going to school across the street so to speak} \\
\multicolumn{2}{l}{\textbf{Topic}: Choosing a college \quad \textbf{Target}: (Bob) um - hum} \\
\midrule
\textbf{ControlVAE-KL-25} &  \textbf{ControlVAE-KL-35}  \\
 \midrule
yeah &  uh - huh   \\
  \midrule
um - hum &  yeah \\
  \midrule
oh that's right um - hum & oh yeah oh absolutely \\
  \midrule
yes & right\\
  \midrule
right & um - hum\\
\midrule
\textbf{Cost annealing (KL=17)} &  \textbf{Cyclical anneal (KL=21.5)}  \\
\midrule
oh yeah &  yeah that's true do you do you do it    \\
 \midrule
uh - huh &  yeah\\
\midrule
right & um - hum\\
\midrule
uh - huh and i think we have to be together & yeah that's a good idea\\
\midrule
oh well that's neat yeah well & yeah i see it too,it's a neat place\\
\bottomrule
\end{tabular}
\end{small}
\end{center}
\end{table}